\documentclass[11pt]{article}

\usepackage[preprint]{acl}
\usepackage{macros}

\usepackage{times}
\usepackage{latexsym}

\usepackage[T1]{fontenc}
\usepackage[utf8]{inputenc}

\usepackage{microtype}

\usepackage{inconsolata}

\usepackage{graphicx}

\title{Task-Centric Acceleration of Small-Language Models}

\author{Dor Tsur\thanks{Work was done during an internship at Amazon.} \\
  \texttt{dortsur93@gmail.com} \\\And
  Sharon Adar\\
  Amazon \\
  \texttt{sharonaz@amazon.com} \\\And
  Ran Levy \\
  Amazon \\
  \texttt{ranlevy@amazon.com} \\}

\begin{document}
\maketitle
\begin{abstract}
Small language models (SLMs) have emerged as efficient alternatives to large language models for task-specific applications. However, they are often employed in high-volume, low-latency settings, where efficiency is crucial.
We propose $\name$, Task-Adaptive Sequence Compression, a framework for SLM acceleration comprising two use-cases: When performing SLM fine-tuning, we propose $\nametok$, which iteratively enriches the tokenizer vocabulary with high-frequency output $n$-grams and then fine-tunes the model to utilize the expanded vocabulary. 
Next, we propose an inference-time method, termed $\namedraft$. $\namedraft$ is a lightweight, training-free speculative decoding method that constructs an $n$-gram draft model from the task's output corpus, mixing task and context $n$-gram information.
$\namedraft$ avoids any additional training, while avoiding draft-target vocabulary alignment constraints. 
We demonstrate the effectiveness of both methods across multiple low output-variability generation tasks.
Our methods show consistent improvements in inference efficiency while maintaining task performance. 
\end{abstract}

\section{Introduction}
Small language models (SLMs) are lightweight fast models that recently rose as a viable alternative to large language models (LLMs) \cite{belcak2025small,subramanian2025small}.
SLMs are often distilled versions of LLMs that pose specific capabilities, with the intention to operate as a part of a multi-agent system \cite{lee2025mapcoder,sharma2025small} or in highly memory-bound high-latency settings, such as on-device deployment \cite{kementchedjhieva2023exploration,kandala2024tinyllm,lu2025demystifying}.
SLMs, while lighter than their larger counterparts, have demonstrated comparable performance when fine-tuned (FT) on specific tasks, from text classification
\cite{luo2023exploring,sun2023text,sakai2025large} and summarization \cite{xu2025evaluating, bailly2025divide}, to  question-answering (QA) \cite{pappas2022data,kweon2024publicly} and code generation \cite{luo2023wizardcoder,pinnaparaju2024stable}.

We argue that tasks naturally suited for SLM FT also often exhibit lower output variability compared to open-ended multi-purpose text generation. 
This manifests in a decrease of the $n$-gram empirical entropy in the dataset of expected task outputs.
That is, expected output texts tend to contain more $n$-grams that repeat across different samples.
These $n$-grams often represent common phrases and terms in the task's expected outputs. 
We analyze this phenomenon through the lenses of typicality \cite[Chapter~2]{cover1999elements}.
% We are interested of leveraging the existence of the typical set of $n$-gram to allow for further efficiency gains in SLMs. \rl{remove last sentence}

\begin{figure*}[!t]
  \centering
  \includegraphics[width=0.9\textwidth, trim=28 150 30 155,
  clip]{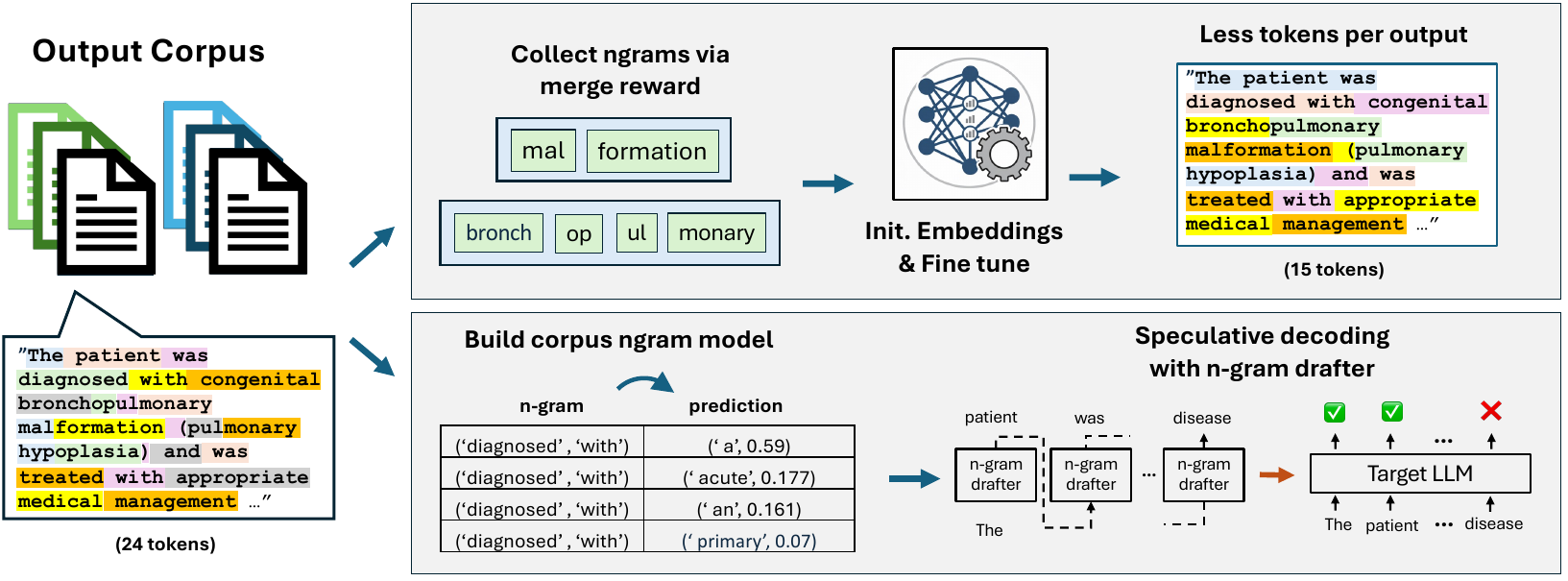}
  \caption{Proposed method for a given task corpus. $\nametok$ (upper) augments the model's tokenizer with $n$-gram tokens and trains their embeddings during the model's fine-tune. $\namedraft$ (lower) accelerates the model in inference-time using an $n$-gram model drafter, providing significant speedup with no additional training. }
  \label{fig:system_fig}
\end{figure*}

% \sa{This parts starts the related work, but it doesn't continue here but in section2. that felt cut in the middle for me.}
To accelerate LLMs, most prior work focus on either decreasing memory requirements or accelerating input processing, with the purpose of handling longer contexts.
Popular methods in these direction are model quantization \cite{lang2024comprehensive,lin2024awq}, KV caching \cite{pope2023efficiently,li2024survey} and attention optimization \cite{dao2022flashattention,kwon2023efficient}.
On the contrary, in this work we follow the low output variability, common in FT SLMs, and focus on the model's outputs. Specifically, we view efficiency in terms of decreasing the total number of model auto-regressive steps required to generate a response from a given task input, which is a known bottleneck in autoregressive generation \cite{openai_latency_optimization_2025}.
To that end, our framework comprising two solutions, which we term Task-Adaptive Sequence Compression ($\name$). Each solution targets a different operational SLM use-case. Figure \ref{fig:system_fig} outlines $\name$'s pipeline.

First, when task-specific fine-tuning of the SLM is performed, we propose $\nametok$, a tokenizer augmentation procedure that explicitly leverages the task-induced concentration of output $n$-grams.
Given a train corpus of expected task outputs, we identify high-frequency $n$-grams whose merge into single tokens yields the largest reduction in expected output length across the corpus. We iteratively enrich the model’s tokenizer with such task-specific tokens using a byte pair encoding (BPE) inspired scheme, and fine-tune the SLM to properly utilize the expanded vocabulary. This procedure reduces the number of decoding steps required to generate typical task outputs, resulting in substantial latency improvements while preserving task performance.

Second, when model fine-tuning is not performed (e.g., when using off-the-shelf fine-tuned SLMs for rapid development) we propose a training-free inference-time acceleration method. Inspired by speculative decoding \cite{leviathan2023fast}, we construct a lightweight $n$-gram draft model directly from the task’s output corpus. The $n$-gram drafter generates sequence predictions at a negligible cost, which are then verified by the target model in a single forward pass.
To balance global task structure with instance-specific information, we combine global corpus-derived $n$-grams with local prompt-derived information into a unified draft model, which we term $\namedraft$. This approach yields meaningful speedups without additional training.
Furthermore, while most existing methods require both the target and draft to share the same vocabulary -- limiting the cross-model applicability of speculative decoding -- $\namedraft$ automatically adapts to the target's vocabulary.

Our main contributions are as follows:
\begin{enumerate}
    \item We characterize the set of generation tasks by the relation between their input vs. output variability, which we then formalize through the information-theoretic perspective of typical $n$-grams.
    % We identify low output variability in a class of generation tasks and formalize this phenomenon through an information-theoretic perspective of typical $n$-grams.
    \item We propose $\nametok$, an algorithm that enriches the SLM's tokenizer vocabulary with task-specific $n$-grams. $\nametok$ integrates seamlessly into parameter-efficient FT, achieving up to $\times 2.1$ reductions in model runtime while preserving generation quality. 
    \item We show that the token distribution's 2-R\'enyi entropy can act as a predictive measure of fine-tuned model acceleration under $\nametok$.
    \item We introduce $\namedraft$, a lightweight, training-free speculative decoding method that uses $n$-gram draft models. $\namedraft$ exhibits inference speedups of up to $\times3.15$ across multiple generation tasks.
\end{enumerate}

\begin{table*}[!t]
\centering
\small
\setlength{\tabcolsep}{7.5pt} % Adjust padding for compactness
\begin{tabular}{| l || c | c | c || c | c | c || }
\toprule
& \multicolumn{3}{|c||}{Bigram Entropy} &  \multicolumn{3}{|c||}{\# Unique Bigrams For $80\%$ CDF}\\
\textbf{Task} & Input & Output & \textbf{$\Delta H$} & Input & Output & Ratio\\
\midrule
\rowcolor{red!15} Sentiment Analysis (Finance) & $14.5$ & $2.21$ & $\downarrow -84.7\%$ & $33.53K$ & $4$ & $8383$\\
\rowcolor{red!15} Yes/No Questions (Medical) & $15.34$ & $3$ & $\downarrow -80.48\%$ & $17.01K$ & $3$ & $5698$\\
\rowcolor{cyan!15} POS Tagging & $15.53$ & $6.57$ & $\downarrow -57.7\%$ & $65.3K$ & $63$ & $1036.3$\\
\rowcolor{cyan!15} Intent Classification \& Slot Fill & $13.4$ & $9.74$ & $ \downarrow-37.45\%$ & $15.943K$ & $774$ & $20.6$\\
\rowcolor{cyan!15} Massive Classification (Legal) & $14.43$ & $12.76$ & $ \downarrow-11.6\%$ & $36.8K$ &  $8.34K$& $4.41$\\
\rowcolor{cyan!15} Abbreviation Expansion (Medical) & $16.17$ & $14.19$ & $ \downarrow-12.24\%$ & $195.1K$ & $84.28K$ & $2.32$\\
\rowcolor{cyan!15}Summarization (Medical) & $16.14$ & $15.39$ & $\downarrow -4.86\%$ & $17.14K$ & $15.16K$ & $1.13$ \\
\rowcolor{cyan!15}Summarization (News) & $17.71$ & $15.79$ & $\downarrow -10.8\%$ & $387.7K$ & $59.34K$ & $6.52$ \\
\rowcolor{cyan!15}Summarization (Legal) & $16.8$ & $15.8$ & $\downarrow -5.9\%$ & $229.9K$ & $89.14K$ & $2.73$ \\
% \rowcolor{cyan!15} Intent Classification \& Slot Fill & $13.4$ & $9.74$ & $ \downarrow-37.45\%$ & $15.94K$ & $774$ & $20.59$\\
\rowcolor{cyan!15}Question Answering (Medical) & $16.2$ & $15.38$ & $\downarrow -5.04\%$ & $196.34K$ & $111.6K$ & $1.76$ \\
\rowcolor{red!15} Creative Writing & $12.75$ & $16.62$ & $\uparrow +30.4\%$ & $11.16K$ & $168.81K$ & $0.06$ \\
\bottomrule

\end{tabular}
\caption{Word bigram distribution analysis across several generation tasks. We compare the Shannon entropy and the number of unique word bigrams that cover $80\%$ of the bigram empirical distribution on an empirical dataset. The specific dataset names for each task is given in Appendix \ref{appdx:implementation}.}
\label{tab:dataset_ent_comparison}
\end{table*}

\section{Related Work}

\textbf{Tokenizer augmentation}
Popularized by \cite{hong2021avocado,sachidananda2021efficient}, modification of the tokenizer's vocabulary had been recently adopted as a method for improving the model's domain-adaptation abilities and sequence compression capabilities.
\cite{gee2023multi} shows that adding the top-$K$ most frequent $n$-grams improved input sequence compression for BERT-like models and \cite{schmidt2025boundless, liu2025superbpe} goes beyond pre-tokenization boundaries to include bigger token merges.
The most relevant work is AdaptiVocab \cite{nakash2025adaptivocab}, which proposes a domain-based vocabulary adaptation method. Domain adaptation targets the input corpus, thus requiring a significant modification of the tokenizer vocabulary. 
In contrast, our setting is \textit{task-driven}: while inputs are linguistically diverse, the task induces low output variability. This asymmetry leads to different design choices. Our method adds a small number of task-specific tokens, capturing most of the relevant variability without sacrificing generality or requiring token removal, and explicitly targeting inference efficiency rather than broad domain coverage.

\textbf{Speculative Decoding:}
Popularized by \cite{leviathan2023fast,chen2023accelerating}, speculative decoding is nowadays a common method for acceleration of LLMs using a fast auxiliary draft model. 
The authors of \cite{fu2024break} study draft parallel decoding using Jacobi iterations. In \cite{cai2024medusa}, a multi-head drafter is proposed, where each head learns to predict a different lookahead step. The authors of \cite{li2024eagle} transform the speculative decoding problem to the feature space via EAGLE.
 \cite{sun2023spectr} shows that speculative decoding is theoretically optimal using optimal transport theory. 
Most related work in this field is $n$-grammys \cite{stewart2024n}, in which a simple bigram model that assigns each token its most probable next token under the given model. However, $n$-grammys is not task-specific, trading generality with performance.

\textbf{$n$-Grams for LLMs:}
Despite the dominance of transformer-based models, $n$-gram statistics remain relevant for efficiency, adaptation, and analysis. Early works combined neural models with external $n$-gram LMs to inject domain-specific biases without retraining \cite{gulcehre2015using}, while later retrieval-based methods generalized $n$-gram memorization through nearest-neighbor interpolation \cite{khandelwal2019generalization}. In a related direction, Infini-grams \cite{liu2024infini} leverage unbounded-length $n$-gram statistics to analyze and approximate LLM next-token behavior. In \cite{diao2021taming} $n$-gram information is incorporated as explicit features or auxiliary representations. Recent analyses further suggest that LLMs often behave similarly to high-order $n$-gram models over substantial portions of the token distribution \cite{nguyen2024understanding,varre2025learning}.

\section{Task-Specific Output Structure and Efficiency }
\label{sec:output_ngrams}
While SLMs are capable of open-ended generation, many practical applications rely on \textit{constrained generation tasks}. In these scenarios, the output is not free-form but strictly bound by a predefined label set, structural schema, or other specific requirements.
In this section, we formalize and quantify this observation through an information-theoretic perspective.

\subsection{Task-Induced $n$-Grams Concentration} 
Consider some generation task $\tau$ that has a representative corpus, i.e., a collection of input texts and corresponding expected outputs. We denote the set of expected outputs for a given task $\tau$ by $\ctask$.
For some finite $n$, we consider the empirical output $n$-gram distribution based on the $n$-gram frequencies in $\ctask$. \footnote{As we are yet to consider a specific tokenizer, we can, without loss of generality, consider word $n$-grams.}
This is a conditional distribution, where the conditioning is on $\tau$.
For example, in POS-tagging, expected output are sequences of discrete part-of-speech labels (e.g., \texttt{DET}, \texttt{NOUN}, \texttt{VERB}), drawn from a small, fixed tag set.
Here, the $n$-gram distribution is significantly skewed towards specific label sequences, unlike the heavier-tailed distribution of open-ended text.

Formally, the skewed $n$-gram distribution exhibits reduced entropy:
\begin{equation}
  H(T^n | \sT=\tau) \ll H(T^n),
\end{equation}
where $T^n\triangleq(T_1,\dots T_n)$ is the $n$-gram random variable and $H(T^n)$ is its entropy, unconditioned on any specific $\tau$.\footnote{Formally, we can take a uniform prior over possible tasks and define $H(T^n)\triangleq |\cT|^{-1}\sum_{\tau\in\cT}H(T^n | \sT=\tau)$.} 
Following the information theoretic notion of \textit{typicality} \cite{shannon1948mathematical,cover1999elements}, the size of the set containing most probable task output $n$-grams, denoted $\mathcal{A}_{n,\tau}$, is determined by this conditional entropy:
\begin{equation}\label{eq:typical_set_size}
  \bigl|\mathcal{A}_{n,\tau}\bigr| \approx \exp\left(H(T^n | \sT=\tau)\right).
\end{equation}
We call $\mathcal{A}_{n,\st}$ the \textit{typical set} of output $n$-grams given generation task $\tau$. 
$\mathcal{A}_{n,\st}$ is potentially exponentially smaller than the full space of $n$-grams. Consequently, a relatively small number of high-probability $n$-grams dominate the task-conditioned distribution. 
% While we consider low \textit{output} variability, we assume that the task's input is given in free-form natural language.

We demonstrate this phenomenon across a range of tasks, using word bigrams for simplicity. 
For each dataset, we measure the bigram entropy and the number of unique bigrams that attain $80\%$ of the bigram CDF.
Both metrics serve as proxies for the size and dominance of the typical set of bigrams in $\ctask$.
As shown in Table \ref{tab:dataset_ent_comparison}, some generation tasks result in a small set of bigrams that cover a large portion of the bigram distribution. 
We examine output variability in a \textit{comparative} manner against the inputs, assuming inputs are provided in free-form natural language.\footnote{We make this assumption as often those may contain human-written texts.}

Notably, the tasks at hand exist along a spectrum, defined in terms of the input vs. output variability. 
To capture this, we measure the difference in entropy and CDF coverage ratio between the input and output bigram distributions.
In highly constrained tasks, such as sentiment analysis, the typical set is very small, implying that most of the output distribution is concentrated on a small set. Conversely, in tasks such as creative writing, outputs exhibit higher variability than inputs.

The comparative quantification distinguishes \textit{tasks} from \textit{domains}; while fixing a domain may reduce variability, this reduction applies to both inputs and outputs, and not necessarily in the same magnitude as outputs may have their unique emphasis or style. 
In contrast, as shown in Table \ref{tab:dataset_ent_comparison}, fixing a task primarily affects the output distribution. 
Consequently, the dominant factor influencing the difference in input vs. output variability is the type of task, rather than the domain.
For example, Yes/No QA exhibits a significant reduction in output variability compared to summarization, although both tasks operate within the medical texts domain.
In what follows, we propose $\name$, which aims to accelerate SLMs whose target is low output variability tasks (except for simple classification tasks that have extremely low variability and don't require acceleration).  As we later explain, each task is aimed at a different use-case and aims to address a different regime of the variability spectrum.

\section{Tokenizer Augmentation Method}\label{sec:sol_tokenizer}
When fine-tuning on the task at hand is both desired and feasible, we extend the standard fine-tuning procedure with a tokenizer preprocessing step.
Specifically, we define a score for merging an $n$-gram of tokens into a single new token. The score is used in a BPE-inspired algorithm to iteratively expand the tokenizer vocabulary with high-scoring $n$-grams. The new $n$-gram token embeddings are optimized alongside the fine tuning procedure.

\begin{figure}[!t]
    \centering
    \includegraphics[width=\linewidth]{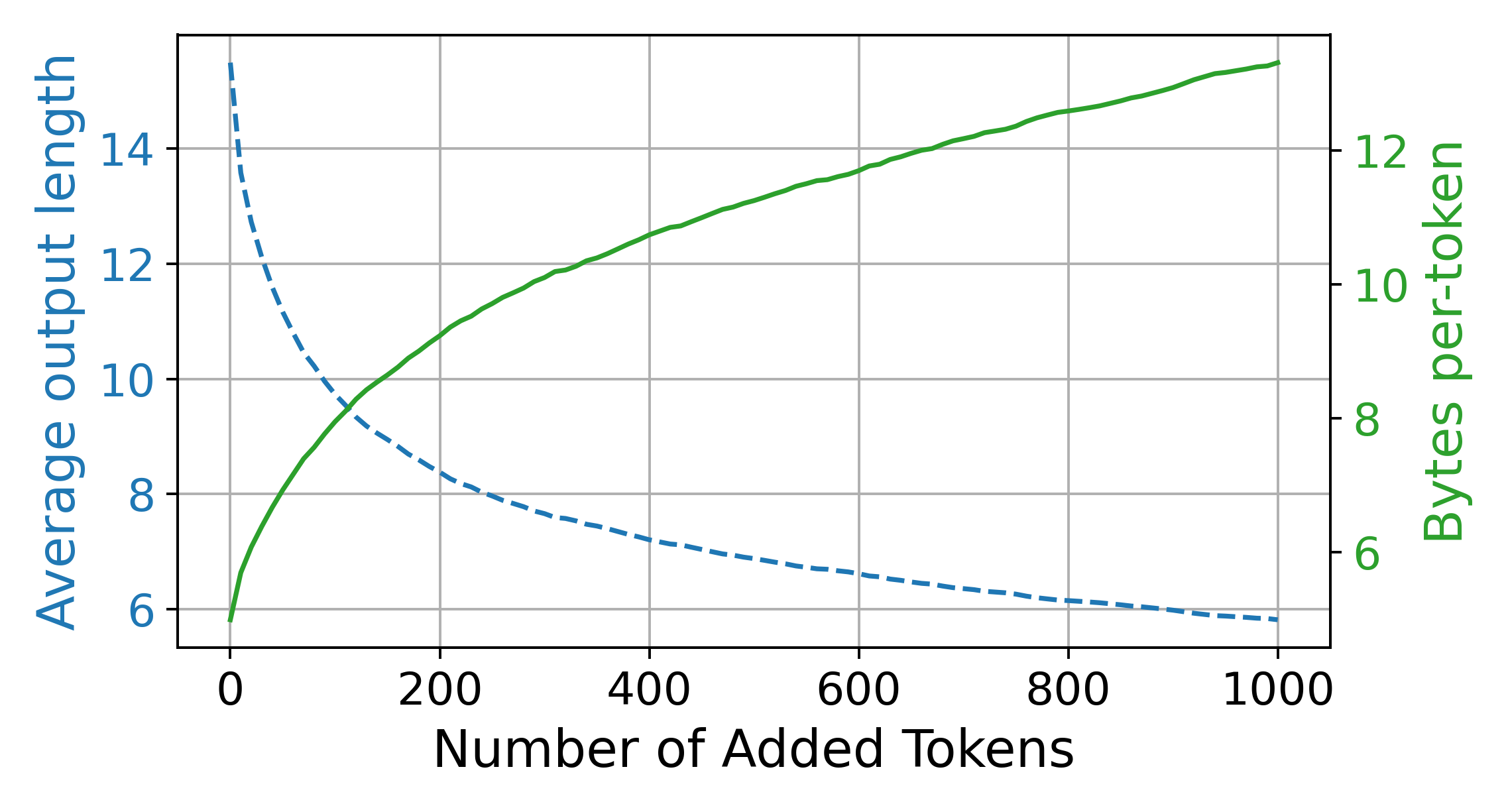}
    \caption{Effect of added tokens on the tokenized output texts in the \text{\eurlex} dataset, $n_{\mathsf{max}}=4$.}
    \label{fig:tokznier_effet}
    % \vspace{-0.5cm}
\end{figure}

\subsection{Enriching the Tokenizer}
Let $\basevoc$ denote the model's vocabulary and let $\ctask$ be the outputs of the train set. We treat it as a representative corpus of the task's outputs distribution, consisting of token sequences over $\basevoc$. 
Notably, $\ctask$ contains only expected outputs, as our goal is to reduce decoding-time latency rather than prefill cost.
We propose a BPE-inspired scheme\footnote{See Appendix \ref{appdx:bpe} for information on BPE.} that iteratively adds token $n$-grams to $\basevoc$ with the purpose of reducing the expected output length in $\ctask$.
To identify potential token $n$-gram candidates, we define the \textit{merge reward} of each token $n$-gram $t^n$ by its expected contribution to sequence-length reduction:
\begin{equation}
f_{\mathsf{merge}}(t^n) \triangleq \mathsf{freq}(t^n)\cdot (n-1),    
\end{equation}
where $\mathsf{freq}(t^n)$ is the $n$-gram's frequency in $\ctask$ and the factor $(n-1)$ corresponds to the net reduction in token count, if we were to merge a single occurrence of $t^n$ into a single token.

While the merge reward captures expected sequence-length reduction, introducing a new $n$-gram token can bias greedy decoding by collapsing multiple plausible continuations into a single dominant path, suppressing alternatives whose prefix collides with the new token's suffix. For instance, adding the bigram ('\texttt{in}', '\texttt{the}') as a new token may lead the model to favor this continuation over individually less frequent but collectively probable alternatives like '\texttt{therapy}' or '\texttt{theorem}', which share the prefix '\texttt{the}'.
To mitigate this effect, we propose the Prefix Collision Score (PCS), which measures the probability that the suffix of an $n$-gram collides with prefixes of existing tokens. Formally, for an $n$-gram $t^n$, let $\mathsf{str}(t^n)$ be its string representation. We define:
$$
\mathsf{PCS}(t^n) = P\left(\bigcup_{t'\in\cV} \mathsf{str}(t_n)\text{ is a prefix of }(\mathsf{str}(t')\right)
$$
for $t'\in\cV_i$ over $\ctask$.
We therefore add $t^n$ to $\mathcal V_i$ only if $\mathsf{PCS}(t^n)$ falls below a prescribed threshold.

Starting from $\cV_0 = \basevoc$, at each step $i$ we extend $\cV_i$ with the element that maximizes $f_{\mathsf{merge}}(t^n)$ and that does not pass the PCS threshold.
This repeats until a predefined vocabulary budget is reached or a predetermined number of iterations are performed.
We refer to this method as $\nametok$ and outline its steps in Algorithm \ref{alg:tokenizer}.
Figure \ref{fig:tokznier_effet} demonstrates the effect of $\nametok$ on \text{\eurlex}, a massive multi-label classification dataset \cite{chalkidis2019large}, in which every text has up to $26$ labels out of $4K$ classes. We evaluate the average output length and measure the output compression rate over the tokenized validation set. Consequently, with only $1000$ new tokens we obtain $\times 2.65$ reduction in the average output length and $\times 2.67$ increase in the bytes-per token metric.

\begin{algorithm}[!t]
\caption{$\nametok$: Tokenizer Enrichment}
\begin{algorithmic}[1]
\Require $\ctask$, $\basevoc$, token budget $M$, $n_{\mathsf{max}}$, PCS threshold $\alpha_{\mathsf{PCS}}$  
\State Initialize $i=0$, $\texttt{usedNgrams} = \texttt{set}()$
\State Set $\cV_i=\basevoc$
\While{$i<M$}
    \State $\texttt{Ngrams}=\texttt{collectNgrams}(\cV_i,n_{\mathsf{max}})$
    \State Calculate $f_{\mathsf{merge}}$ for $t^n\in\texttt{Ngrams}$
    \State Take $t^\star\in\argmax_{t^n\in\cV_i^n}f_{\mathsf{merge}}(t^n)$
    \If{$\mathsf{PCS}(t^\star) < \alpha_{\mathsf{PCS}} $}
        \State $\cV_i \leftarrow \cV_i \cup (t^\star)$
        \State $i \leftarrow i+1$
    \EndIf
    \State $\texttt{usedNgrams} \leftarrow t^\star$   
%%%%%
\EndWhile
\State \Return Enriched tokenizer vocabulary $\cV$
\end{algorithmic}
\label{alg:tokenizer}
\end{algorithm}

We find additional motivation in the uniform information distribution (UID) hypothesis~\cite{jaeger2006speakers}, which posits that efficient linguistic representations tend to distribute information evenly across tokens, leading to approximately uniform per-token surprisal along sequences. 
This principle motivates our procedure. By collapsing typical high-probability $n$-grams into single tokens, the induced token distribution over the enriched vocabulary becomes less peaked, increasing the token distribution entropy and expanding the typical set. For example, on the \text{\eurlex} dataset, enriching the tokenizer with $1000$ task-specific $n$-grams increases the normalized entropy of the empirical output token distribution from $0.423$ to $0.592$ (see Appendix~\ref{appdx:uid} for more information).

\subsection{Integration with Model Fine-Tuning}
Equipped with the extended tokenizer, we integrate the new tokens into the model's architecture. 
This requires resizing the model's input embedding matrix and the LM head to accommodate the new $n$-gram tokens.
For a new $n$-gram token $t_{\mathsf{new}}=(t_1,\dots,t_n)$, we initialize its embedding vector as the mean of its constituent tokens embeddings
$\mathbf{e}_{\mathsf{new}} = \frac{1}{n} \sum_{j=1}^{n} \mathbf{e}_{t_j}$.
This initialization provides a strong semantic prior, ensuring stable training with minimal adaptation overhead.
We apply this initialization strategy to both the input embeddings and the LM head weights.

During the task-specific fine-tuning phase, these new parameters are optimized jointly with the model's low-rank adapters. 
Importantly, the additional training overhead remains modest and is compatible with single-GPU fine-tuning regimes commonly associated with SLM deployment.
For example, when fine-tuning a Qwen2.5-3B model with $r=16$, adding $500$ new tokens results in an increase of the total trainable parameters by $6\%$.
As shown in Section \ref{sec:results}, this integration yields significant efficiency gains in inference latency while maintaining output generation quality.

\section{$n$-Gram Speculative Decoding Method}\label{sec:sol_specdec}
When fine tuning is not intended or a fine-tuned model is already available,
we can still use the ideas from Section \ref{sec:output_ngrams} to accelerate the model under the given task.
We follow a speculative decoding procedure building upon the typical set of $n$-grams
in $\ctask$.
The resulting method is a training-free inference-time acceleration algorithm that is model agnostic.

\subsection{Speculative Decoding}
Speculative decoding \cite{leviathan2023fast} is a recent inference acceleration methodology for LLMs in memory-bound environments, which follows a `\textit{draft, then verify}' scheme. 
In speculative decoding, a fast draft model generates a sequence of $\gamma$ tokens, which are then verified by the slower target model in parallel via a single forward pass. Following verification, a subsequence of $k \leq \gamma$ tokens is accepted.
The key trade-off in speculative decoding lies between the target's acceptance rate of draft tokens and the draft generation time.
Conceptually, faster drafts typically approximate the target distribution less accurately, while increasing acceptance rates often comes at the price of slowing the draft model.

\subsection{Proposed Method: $\namedraft$}
Our goal is to construct an $n$-gram draft model that captures the typical $n$-gram set in $\ctask$. First, fix $n_{\mathsf{max}}\in\mathbb{N}$. The drafter is a collection of $n$-gram models $\sM_n$ for $n\leq n_{\mathsf{max}}$. Each $\mathsf{M}_n$ is derived from the $n$-gram frequencies in $\ctask$.
To retain only significant $n$-grams, we prune each $\sM_n$ by removing elements whose frequency is below a prescribed threshold $p_{\mathsf{min}}$.
Consequently, we define the \textit{corpus} $n$-gram drafter as $p_{\mathsf{corp}}\triangleq \bigcup_{n\leq n_{\mathsf{max}}}\mathsf{M}_n$. Given a prefix $c = t^{n_{\mathsf{max}}}$ we obtain $p_{\mathsf{corp}}(\cdot|c)$ by querying $(\mathsf{M}_n)_{n\leq n_{\mathsf{max}}}$ in descending order of $n$. 
We return the first matching prediction for $c$ in $p_{\mathsf{corp}}$, or the most frequent token in $\cV$ if none was found.

\begin{table*}[!t]
    \centering
    \small
    \begin{tabular}{||c|c|c|c|c|c|c||}
    \toprule
        \# Added Tokens & $0$ & $150$ & $300$ & $500$ & $750$ & $1000$    \\
        \midrule
         Judge Score & $3.53 \pm 0.73$ & $3.21 \pm 0.79$ & $3.15 \pm 0.82$ & $3.23 \pm 0.77$ & $3.25 \pm 0.76$ & $3.18 \pm 0.79$ \\
         \bottomrule
    \end{tabular}
    \caption{LLM judge results vs. Number of added tokens, $\nametok$, \text{\clinical} dataset.
    % There is an average decrease of $\approx 0.3$ ($8.5\%$), while the results std is above $0.78$ in all cases.
    }
    \label{tab:judge_llm_asclepius}
\end{table*}

\begin{figure*}[!t]
    \centering
    \begin{subfigure}[b]{0.31\textwidth}
        \centering
        \includegraphics[width=\textwidth, trim=10 10 28 10,clip]{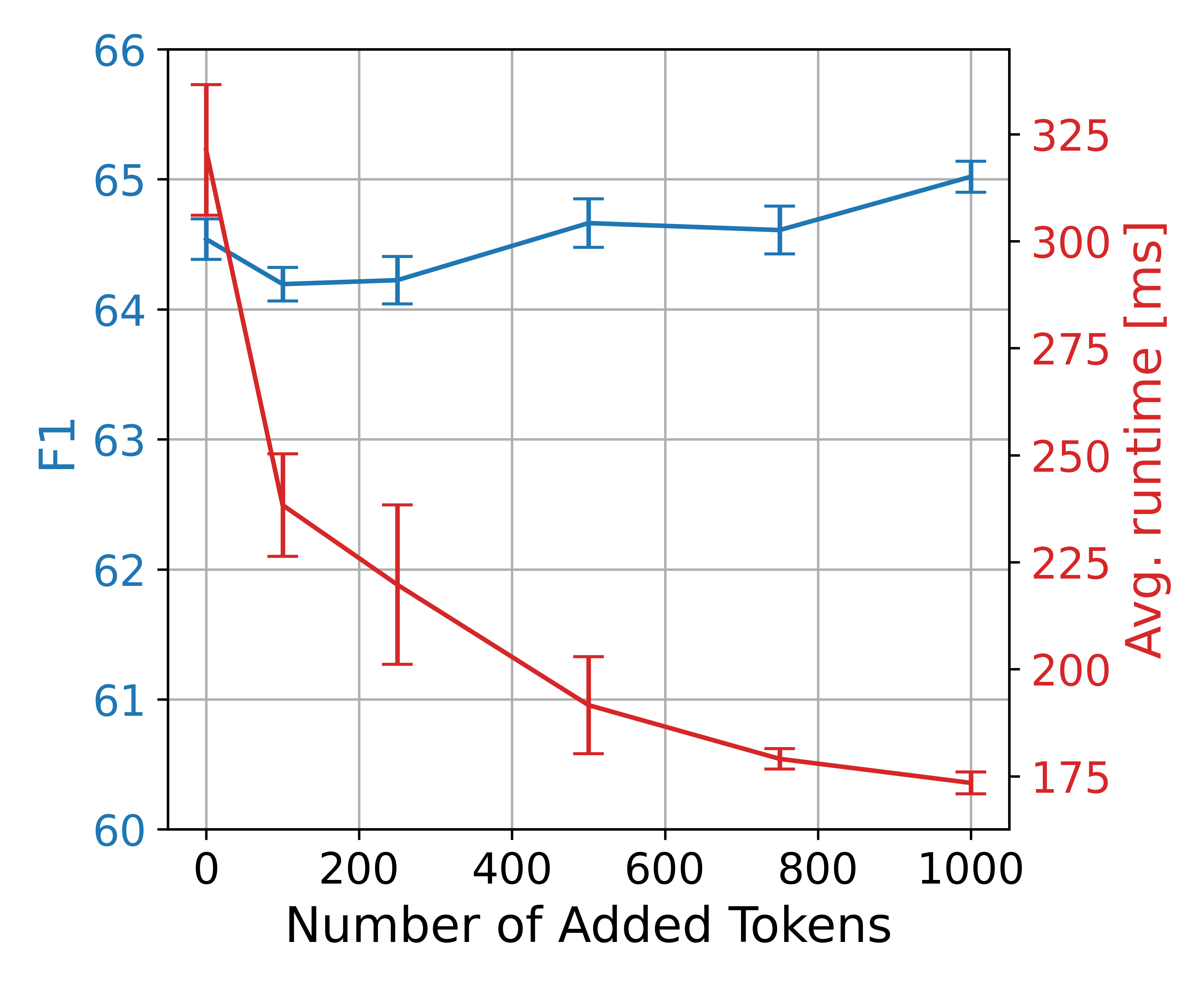}
        \caption{\text{\eurlex} (massive classification)}
        \label{fig:results_eurlex}
    \end{subfigure}
    \hfill
    \begin{subfigure}[b]{0.3\textwidth}
        \centering
        \includegraphics[width=\textwidth, trim=10 10 26.7 10,clip]{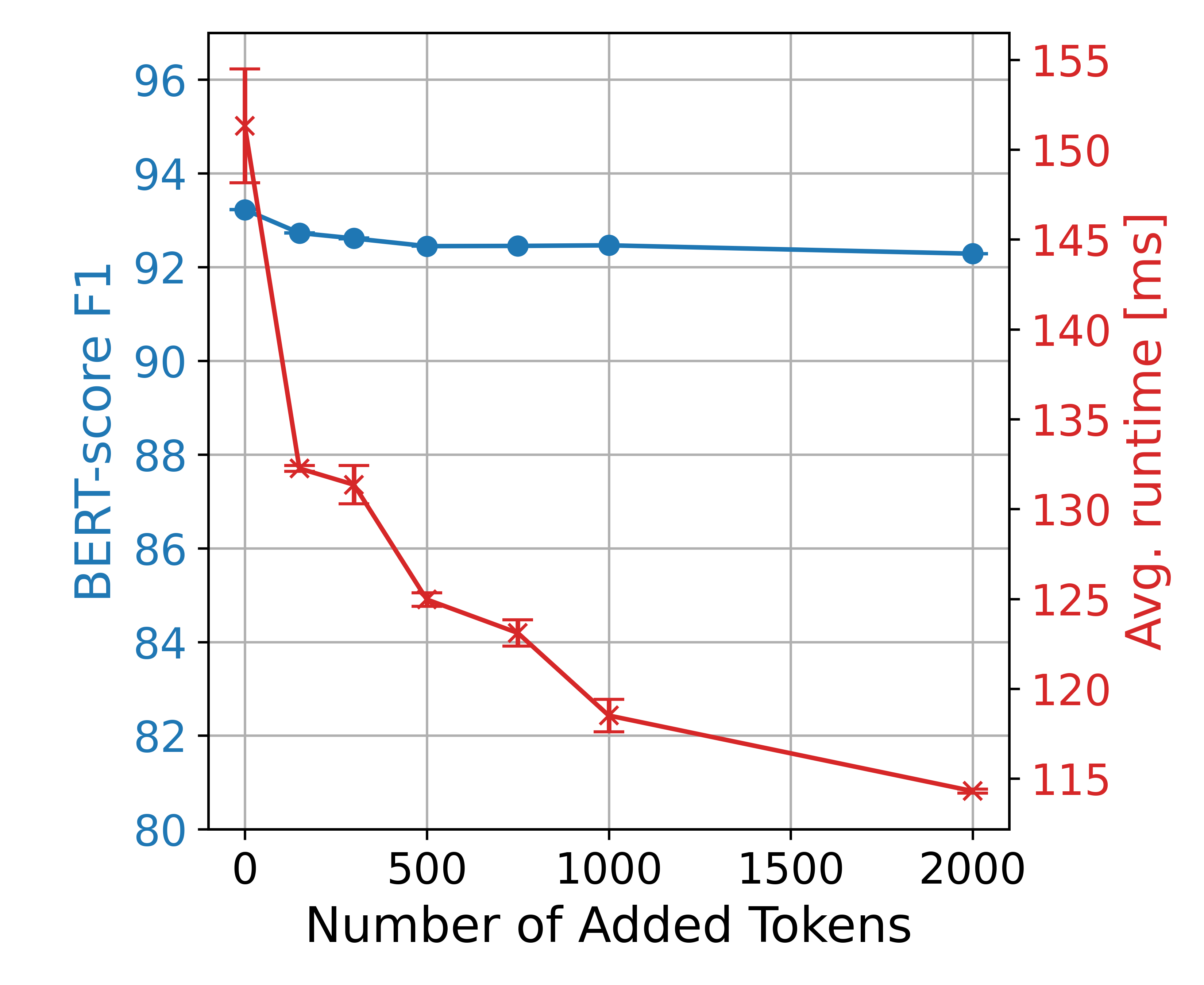}
        \caption{\text{\clinical} (medical QA)}
        \label{fig:results_asclepius}
    \end{subfigure}
    \hfill
    \begin{subfigure}[b]{0.31\textwidth}
        \centering
        \includegraphics[width=\textwidth, trim=10 10 10 10,clip]{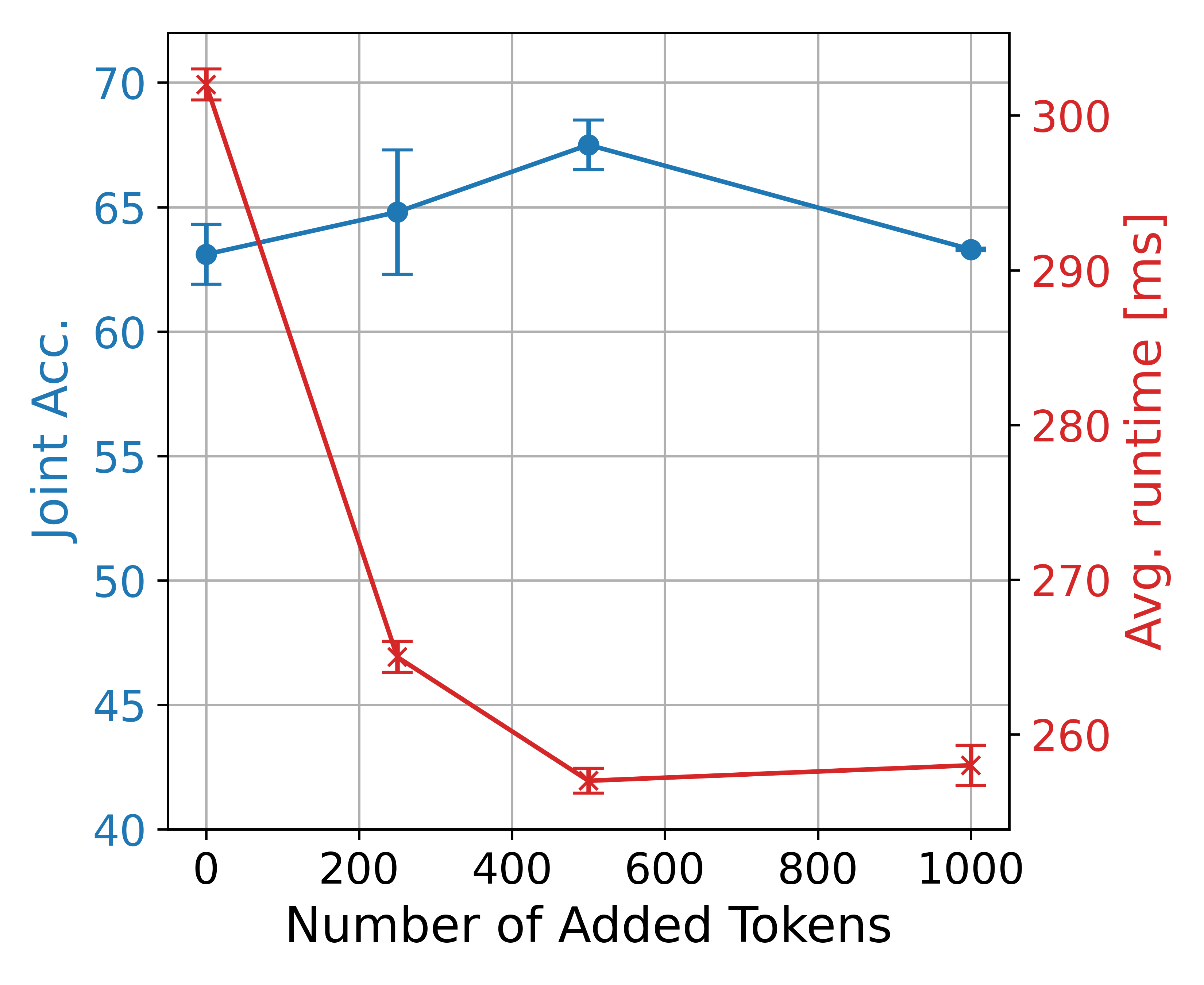}
        \caption{\text{\massive} (ICSF)}
        \label{fig:results_icsf}
    \end{subfigure}
    \caption{$\textbf{\nametok}$: \textcolor{blue}{Task performance metric} vs. \textcolor{red}{average runtime} vs. Number of Added tokens. Each dataset has a different metric according to its task: F1 for \text{\eurlex}, Bert-score for \text{\clinical} and join accurcay for \text{\massive}.}
    \label{fig:main_tokenization_results}
\end{figure*}

Since $p_{\mathsf{corp}}$ is derived solely from $\ctask$, it mostly captures global task information, while context-specific information may be valuable for the next token prediction.
For example, in QA, overloading the previous context term captures specific phrases that are useful for the generation, but are not necessarily in $\ctask$.
We incorporate this local, context-dependent information by combining the corpus $n$-gram with a prompt-based $n$-ngram drafter $p_{\mathsf{prompt}}$. The prompt drafter is obtained similarly to $p_{\mathsf{corp}}$, but it is based on the available context. 
The resulting $n$-gram drafter, termed $\namedraft$, is a mixture distribution. For some $\lambda\in[0,1]$ it is given by 
$$
    p_{\mathsf{mixed}}(x|c) = \lambda p_{\mathsf{corp}}(x|c) + (1-\lambda) p_{\mathsf{prompt}}(x|c).
$$
$\namedraft$ synergies global and local information.

The proposed method alleviates several important bottlenecks of existing speculative decoding methods.
First, it does not require an additional parametric drafter model, thus reducing memory, the main overhead of speculative decoding. 
Furthermore, the method is training-free, i.e., fine-tuning the target model does not require re-training the drafter.
Next, the method only requires that $\ctask$ faithfully represent the target's output distribution  -- a reasonable assumption when $\ctask$ is derived from the target's training data.
Finally, it is tokenizer-agnostic as $\nametok$ operates on the token-level. It is therefore independent of any specific tokenizer choices and tokenizer-matching, which is a current issue in the speculative decoding literature with only preliminary solutions \cite{timor2025accelerating}.

\paragraph{Trade-off between methods.}
As noted in Section \ref{sec:output_ngrams}, tasks can be arranged according to output $n$-grams variability, which is a proxy for typical $n$-gram set's size.
When the typical set is small, $\nametok$ with a reasonable token budget has a higher probability of covering a large portion of it, implying higher potential efficiency gains. 
However, when the set is large, $\nametok$ would require a larger budget to succeed, resulting in a complex fine-tune procedure.
In such cases, $\namedraft$ would be favorable, as it remains a lightweight solution compared to an LLM draft alternative even if it considers a large amount of $n$-grams.

\section{Results}\label{sec:results}
We study the empirical performance of the proposed methods on a set of experiments spanning $4$ datasets. All experiments were performed on a single Tesla V-100 GPU (32GB). Hyperparameter values are based on preliminary evaluations and ablations. Full details are given in Appendix \ref{appdx:implementation}.

\subsection{$\nametok$: Tokenizer Augmentation}
We test the effect of $\nametok$ on the resulting fine-tuned model. 
We measure the effect of the added tokens (and consequently, trainable parameters) on the trained model's performance. Our baseline is the same base model, going through similar training, but without the newly introduced $n$-gram tokens.
% We demonstrate the effectiveness of our method on two tasks within the output variability spectrum: Massive multi-class and multi-label test classification and medical QA. 
In all experiments we consider a Qwen2.5-0.5B base model, employ QLoRa optimization \cite{dettmers2023qlora} and train for $3$ epochs. 
Due to space constraints, additional performance metrics and ablations are reported in Appendix \ref{appdx:more_results}.

% \begin{figure}
%     \centering
%     \includegraphics[width=\linewidth]{figures/f1_runtime_ngram_4.png}
%     \caption{$\nametok$: F1 and average runtime vs. \# added tokens, EUR-LEX dataset, $n_{\mathsf{max}}=4$.}
%     \label{fig:results_asclepius}
% \end{figure}

\noindent\textbf{Datasets.} We consider three tasks: (i) massive text classification for which we use the \text{\eurlex} dataset \cite{chalkidis2019large}. It contains $19.6K$ policy texts, each labeled with up to $26$ labels out of a total of $4K$ possible labels, (ii) Medical QA, for which we consider the \text{\clinical} dataset \cite{kweon2024publicly}. We focus on the QA category that consists of $11K$ QA pairs, (iii) Intent classification \& slot filling (ICSF), for which we consider the \text{\massive} dataset \cite{fitzgerald2023massive} that contains of $16.5K$ english utterances with intent and slot fills labels. The dataset contains $50$ distinct intent categories and $60$ slot fill categories.

\noindent\textbf{Metrics.} For each dataset, we measure the average time to generate a response and calculate the response's quality on a task-specific evaluation metric. For \text{\eurlex} we consider the F1 score, and for \text{\clinical} we consider both the BERT score \cite{zhang2019bertscore} and a LLM-judge score, based on the evaluation prompt from \cite{kweon2024publicly} using Claude4.5-Sonnet as the judge. 
For ICSF we consider both intent accuracy, slot fill F1 score and joint accuracy.
To follow a real-world use case, we consider batched generation with $b=8$.

\noindent\textbf{Results.} 
We examine the effect of the number of tokens added, comparing to the same base model undergoing comparable training without newly introduced $n$-gram tokens.
As shown in Figure \ref{fig:main_tokenization_results}, training with the extended tokenizer preserves task performance while achieving faster average generation time on the test set. 
Table \ref{tab:judge_llm_asclepius} measures a judge-LLM evaluations of the \text{\clinical} outputs, showing negligible impact on the output quality. 

Our experiments consider a token budget of up to $1000$ $n$-grams. While increasing the number of added tokens consistently reduces inference-time generation latency, the marginal gains diminish as the budget grows, indicating a trade-off between acceleration and the training parameters overhead. In practice, the effective cutoff improvement in our experiments is no larger than $500$ tokens, which is significantly smaller than the token budgets typically used in domain-base tokenization adaptations \cite{nakash2025adaptivocab}. We leave a principled analysis of the optimal $\nametok$ budget for future work.

\noindent\textbf{Predicting Runtime Reduction.}
Fine-tuning is computationally expensive, motivating the need for a lightweight proxy to assess a model's potential benefit from $\nametok$. 
We examine whether a dataset-level statistic can serve as a reliable predictor of inference runtime. Specifically, we consider the 2-Rényi entropy ($H_2$) of the token distribution, previously linked to downstream performance.
Our experiments comprise $124$ runs across $8$ configurations, where runs within each configuration differ only in the token budget.
We consider two complementary criteria:
(i) Kendall's $\tau$, computed separately for each configuration, yields $\tau < -0.9$ ($p < 0.005$) in all cases, indicating a strong and consistent inverse correlation.
We measure the rank correlation using Kendall’s $\tau$, all yielding  $\tau < -0.9$ with $p$-value $<0.005$. This indicates a near-perfect inverse agreement.
(ii) Following \cite{pesaran1992simple}, we asses how often a change in $H_2$ correctly predicts the direction of change in runtime. We observed a $92\%$ success rate, confirming that $H_2$ is a highly reliable indicator for predicting runtime reduction.
Additional details are provided in Appendix~\ref{app:proxy_validation}.

\subsection{$\namedraft$: Inference-Time Acceleration}
We compare our results with the n-grammy's method \cite{stewart2024n}, which proposes a bigram-based drafter that is obtained by taking the conditional LLM distribution for each element in the tokenizer's vocabulary.
To maintain comparability we consider task fine-tuned LLMs, to ensure that its conditional distribution is adapted to the task as hand. We consider a beam-size of $b=1$ and apply $\namedraft$ with $p_{\mathsf{min}}=5$ and $\lambda=0.75$. 

\noindent\textbf{Datasets and Models}
We consider two medical QA tasks, given by the \text{\clinical} \cite{kweon2024publicly} and MedQuaD \cite{ben2019question} datasets, and \text{\eurlex}.
\text{\clinical} focuses on reading comprehension of medical texts, whereas MedQuaD does not contain an additional context for each question.

\begin{figure}[!t]
    \centering
    \includegraphics[width=\linewidth]{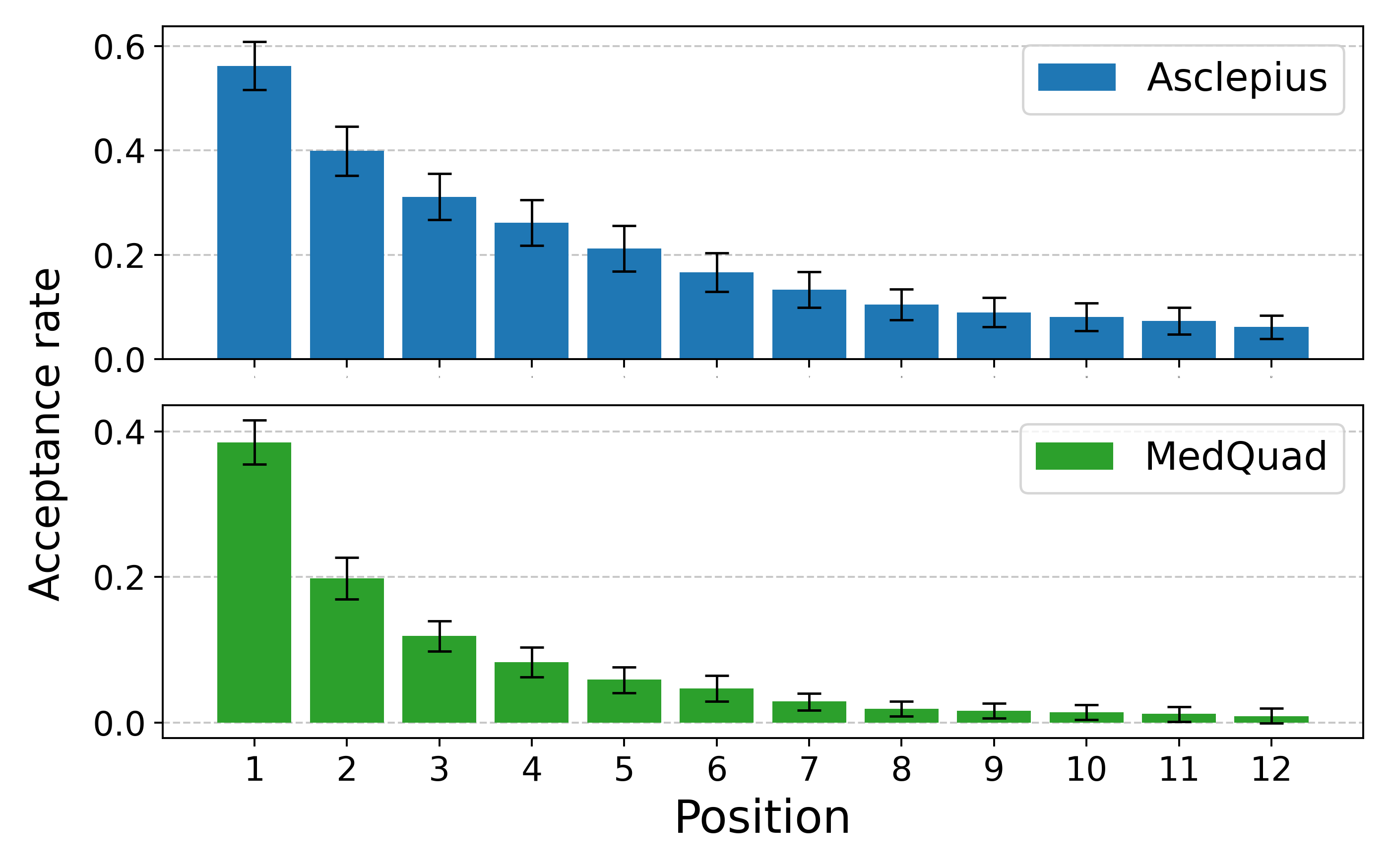}
    \caption{Acceptance rate vs. speculated token position.}
    \label{fig:accept_vs_pos}
    % \vspace{-0.4cm}
\end{figure}

\noindent\textbf{Speedup over target model.}
We evaluate performance by calculating the average speedup over the target model across $50$ test samples per generation task. We report the performance of the corpus $n$-gram $p_{\mathsf{corp}}$, the n-grammy's drafter, and their mixing with the context-based $n$-gram drafter.
As seen in Table \ref{tab:speedup_combined}, the corpus $n$-gram achieves a better target speedup over the n-grammy's drafter. This advantage is maintained after mixing with $p_{\mathsf{prompt}}$, emphasizing the power of global corpus information for speculative decoding.

\noindent\textbf{Position-wise Acceptance.}
Due to the negligible cost of obtaining a prediction from $n$-gram models we take long prediction horizons, which are then verified \textit{in-parallel} by the target model, without suffering a significant latency overhead.
We provide a finer granularity for acceptance by analyzing the acceptance rate for each position in $\namedraft$'s continuations separately. As seen in Figure \ref{fig:accept_vs_pos}, the acceptance rate drops as the position index increases. 
This is due to the expected distribution shift of the $n$-gram model.
Although the $n$-gram model is a highly simplified language model, we obtain relatively high acceptance rates in initial positions, with first position acceptance rate $\alpha=0.57$ in \text{\clinical} and $\alpha=0.39$ in MedQuaD.

\begin{table}[]
    \centering
    \small 
    \begin{tabular}{||l|c|c||}
        \toprule
        \textbf{Dataset} & \textbf{Method} & \textbf{Speedup} \\
        \midrule
        \multirow{4}{*}{\text{\clinical}}
          & $p_{\mathsf{corpus}}$ & $\times1.63$ \\
          & $\namedraft$ \textbf{(ours)} & $\mathbf{\times3.15}$ \\
          & N-Grammys (bigram) & $\times1.03$ \\
          & N-Grammys (mixed) & $\times3.08$ \\
        \midrule
        \multirow{4}{*}{MedQuaD}
          & $p_{\mathsf{corpus}}$ & $\times1.52$ \\
          & $\namedraft$ \textbf{(ours)} & $\mathbf{\times1.78}$ \\
          & N-Grammys (bigram) & $\times1.01$ \\
          & N-Grammys (mixed) & $\times1.32$ \\
        \midrule
        \multirow{4}{*}{\text{\eurlex}}
          & $p_{\mathsf{corpus}}$ & $\times 1.49$ \\
          & $\namedraft$ \textbf{(ours)} & $\mathbf{\times 2.17}$ \\
          & N-Grammys (bigram) & $\times 1.04$ \\
          & N-Grammys (mixed) & $\times 1.98$ \\
        \bottomrule
    \end{tabular}
    \caption{Target speedup results in several datasets. $\namedraft$ is employed with $\lambda=0.75$.}
    \label{tab:speedup_combined}
    % \vspace{-0.4cm}
\end{table}

\section{Conclusion}
This paper presented $\name$, an SLM acceleration framework that exploits the $n$-gram typical set in low output variability tasks.  The proposed methods consider either fine-tune setting via tokenizer augmentation or inference-time acceleration using $n$-gram draft models.
Moreover, we study the power of the token distribution's R\'enyi entropy as a predictive measure of potential runtime acceleration.
The proposed methods substantially reduce the number of decoding steps while preserving output quality.
For future work we consider a comprehensive analysis of the optimal token budget for $\nametok$ following the diminishing returns in terms of newly added tokens.

\section{Limitations}
The proposed methodologies suffer from several limitations. First, we assume that the considered task has an available representative dataset, which we can use for typicality analysis. We implicitly assume that the training dataset we utilize captures the underlying distribution in the test set.
The success of our method is limited to generation tasks that exhibit the low output variability demonstrated in Table \ref{tab:dataset_ent_comparison}. 

The tokenizer augmentation procedure hinges on the assumption that we have access to the model weights and that we have the computational capabilities to fine-tune that model. Furthermore, by augmenting the model's tokenizer, we may harm its generalist capabilities and decrease its performance in other tasks. However, we anticipate this effect to be less significant than both adding and removing tokens from the vocabulary, as often done in domain-based methods (as, e.g., in \cite{nakash2025adaptivocab}).

\bibliography{bibliography}

\clearpage

\appendix

\begin{table*}[!t]
\centering
\small
\setlength{\tabcolsep}{7.5pt} % Adjust padding for compactness
\begin{tabular}{| l | c | }
\toprule
Task & Dataset Name\\
\midrule
Sentiment Analysis (Finance) & \cite{malo2014good}\\
Yes/No Questions (Medical) &  \cite{kweon2024publicly}\\
 POS Tagging & \cite{silveira14gold}\\
 Intent Classification \& Slot Fill &  \cite{fitzgerald2023massive}\\
 Massive Classification (Legal) & \cite{chalkidis2019large}\\
 Abbreviation Expansion (Medical) &  \cite{kweon2024publicly}\\
Summarization (Medical) &  \cite{kweon2024publicly} \\
Summarization (News) & \cite{hermann2015teaching}\\
Summarization (Legal) & \cite{shukla2022legal} \\
Question Answering (Medical) &  \cite{kweon2024publicly} \\
 Creative Writing & \cite{fan2018hierarchical} \\
\bottomrule

\end{tabular}
\caption{Information on the datasets used to generate Table \ref{tab:dataset_ent_comparison}.}
\label{tab:datasets_info}
\end{table*}

\section{Additional Information on Byte-Pair Encoding}\label{appdx:bpe}
BPE \cite{gage1994new} is a simple frequency-based algorithm that constructs a subword vocabulary in a data-driven manner.
Given a training corpus $\mathcal{C}$, BPE learns both a vocabulary $\mathcal{V}$ and an ordered list of merge operations.

The algorithm starts from a base vocabulary $\mathcal{V}_0$, typically consisting of individual characters or bytes.
Each element of the corpus is initially represented as a sequence of symbols from $\mathcal{V}_0$.
At each iteration, BPE identifies the most frequent token bigram in the corpus and merges it into a new token, which is added to the vocabulary.
This merge operation replaces all occurrences of the selected pair with the newly created single token.

The merging procedure iteratively repeats for $M$ iterations, resulting in a final vocabulary
$\mathcal{V} = \mathcal{V}_0 \cup \{t^\star_1, \dots, t^\star_M\}$,
along with an ordered merge table that records the sequence in which merges were applied.
At inference time, new text is tokenized by greedily applying these learned merges in order, producing a sequence of tokens drawn from $\mathcal{V}$.
Following the HuggingFace implementation of tokenizer vocabulary augmentation, when adding new tokens to an existing vocabulary, their corresponding merge operations are not added to the merge table. They are stored in a different data structure. Upon encoding, the set of new tokens is queried for a possible continuation prior to the base tokenizer's merge table.

\section{Additional Implementation Details}\label{appdx:implementation}

\subsection{LLM Judge Information}
The judge we use to generate the results in Table \ref{tab:judge_llm_asclepius} is Claude-4.5 Sonnet. We deploy the LLM evaluation pipeline through Amazon Bedrock. The scores are calculated out of 1260 output samples. The evaluation prompt is given Figure \ref{judge_propt}. It is based the judge evaluation prompt from \cite{kweon2024publicly}. To avoid position bias in the response position in the evaluation prompt, we randomly select the order of LLM response in each sample.

\subsection{Additional Training Information}\label{appdx:taining}
Applying $\nametok$ requires fine-tuning the model at hand with the updated token vocabulary.
We fine-tuned the model using Quantized Low-Rank Adaptation (QLoRA) \cite{dettmers2023qlora} with 4-bit quantization via the \texttt{BitsAndBytes} library, employing the \texttt{NF4} quantization type with double quantization enabled and \texttt{bfloat16} precision for compute. We consider LoRa rank $r=16$ and $\alpha=16$ (i.e., a scaling factor of $1.0$), and dropout rate of $0.2$. The LoRA adapters were applied to the attention layers as well as the feed-forward layers. Additionally, the embedding layer and language modeling head were retrained to accommodate the extended vocabulary with new task-specific tokens. Training was conducted with an initial learning rate of $2\times10^{-4}$ using a linear learning rate scheduler and the \texttt{AdamW} optimizer \cite{loshchilov2017decoupled}. We used a per-device batch size of $8$ and an effective batch size of $64$.

\subsection{Additional Datasets and Models Information}
The datasets considered in Table \ref{tab:dataset_ent_comparison} are described in Table \ref{tab:datasets_info}. In our experiments, we consider the following datasets:
\begin{enumerate}
    \item \text{\clinical} \cite{kweon2024publicly}, a synthetic dataset of discharge summaries and instructions, aimed toward domain-based reading comprehension. It contains 8 generation tasks. In this work we consider the QA task. It comprises $11K$ samples of (discharge summary, instruction) pairs and an expected output. 
    \item MedQuaD \cite{ben2019question}, an open-ended QA dataset, where the questions do no contain any additional context. This dataset tests the model's knowledge rather than its retrieval and context comprehension capabilities. The dataset contains $10K$ samples.
    \item \text{\eurlex} \cite{chalkidis2019large}, a massive multi-label multi-class classification dataset of policy texts. Each sample contains a a collection of EU policy recitals and a sequence of labels. To transform this task into a text generation task, we map each label to its textual name, given by a provided label-name dictionary. To maintain a constant order of labels, we order each label sequence numerically according to label ID prior to its decoding into a sequence of label names. The dataset contains $57K$ samples, each with up to $26$ labels out of a set of $4K$ labels.
    \item \text{\massive} \cite{fitzgerald2023massive} an intent-classification \& slot filling dataset, curated over $52$ languages. Each sample contains an input text, alongside a pair of an expected intent and extracted slot fills. We convert the ICSF task into a generation task by transforming the labels into the string 
    \texttt{f"intent: [INTENT], slot fills: \{ [SLOT FILL 1], \dots, [SLOT FILL M] "}, which serves as the label for task fine-tune. We consider the english language subset of the \text{\massive} dataset.
\end{enumerate}

For speculative decoding, we consider the following fine-tuned models:
\begin{enumerate}
    \item For MedQuaD we consider the QLoRa fine-tuned medical model  \url{https://huggingface.co/ae-aydin/Llama-3-8B-Instruct-Medical-QLoRA}
    \item For \text{\clinical} dataset we consider a fine-tuned Llama model \url{https://huggingface.co/starmpcc/Asclepius-Llama3-8B}
    \item For the \text{\eurlex} dataset we consider a locally fine-tuned Qwen2.5 model that based fine-tuned according to the procedure from Appendix \ref{appdx:taining}.
\end{enumerate}

\begin{figure*}[!t]
    \centering
    \begin{subfigure}[b]{0.45\textwidth}
        \centering
        \includegraphics[width=\textwidth]{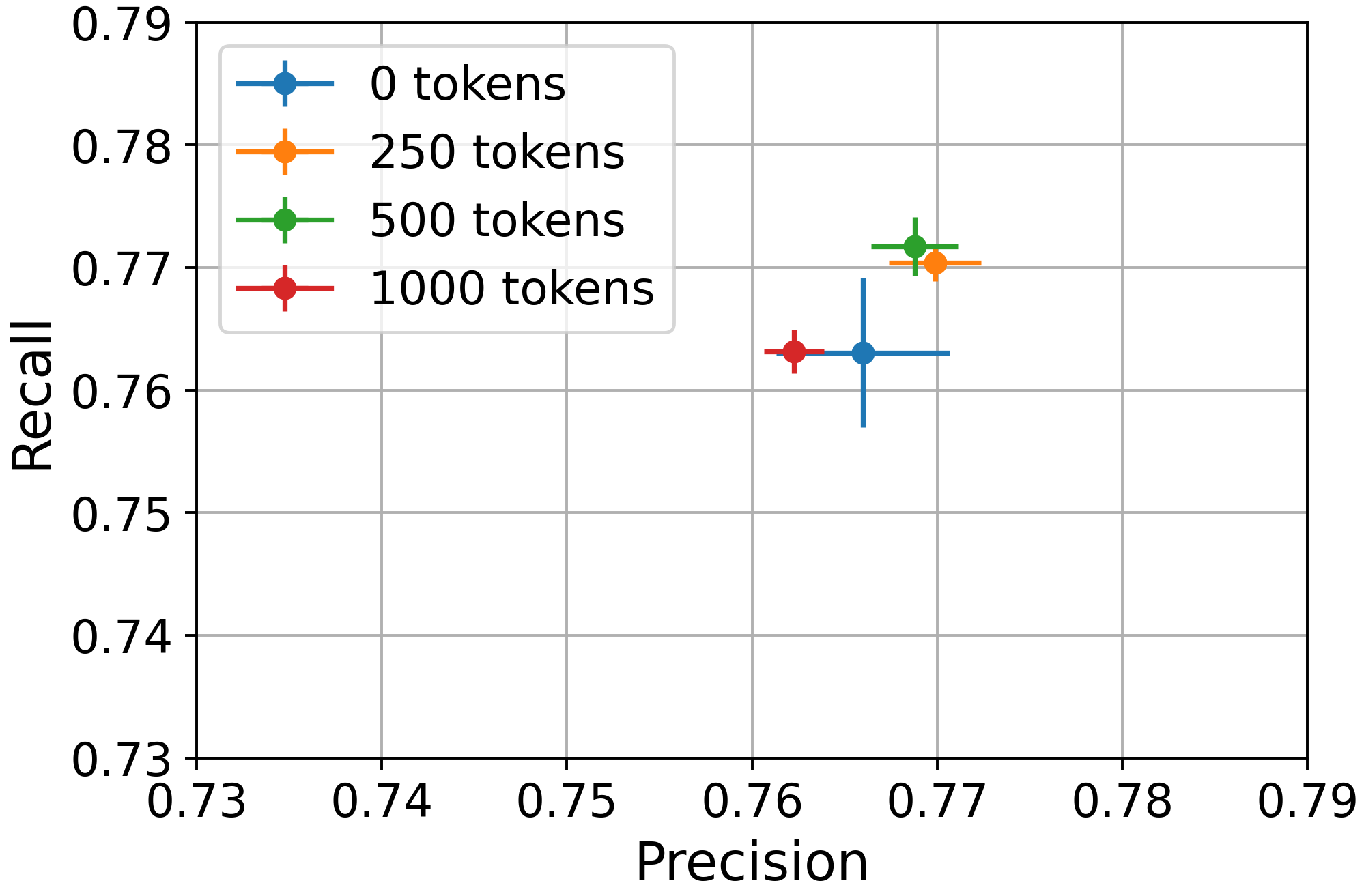}
        \caption{$3$ Training Epochs}
        \label{fig:sub1}
    \end{subfigure}
    \hfill
    \begin{subfigure}[b]{0.45\textwidth}
        \centering
        \includegraphics[width=\textwidth]{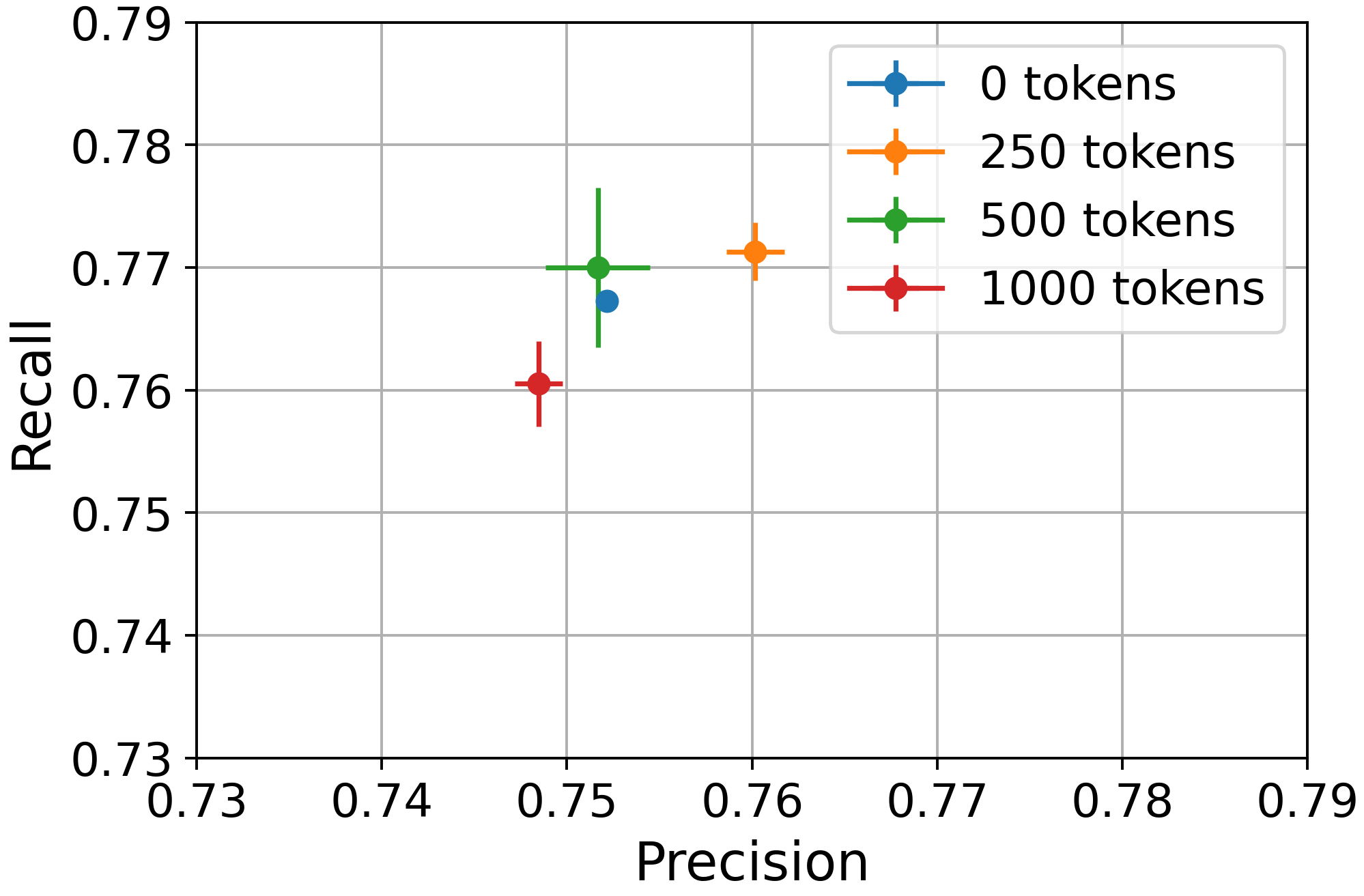}
        \caption{$3$ Training Epochs}
        \label{fig:sub2}
    \end{subfigure}
    \caption{Precision-recall trade off for the ICSF task. Each marker denotes a different number of newly added token $n$-grams to the tokenizer vocabulary prior to fine-tuning.}
    \label{fig:pr_icsf}
\end{figure*}

\begin{figure*}[!t]
    \centering
    \begin{subfigure}[b]{0.32\textwidth}
        \centering
        \includegraphics[width=\textwidth]{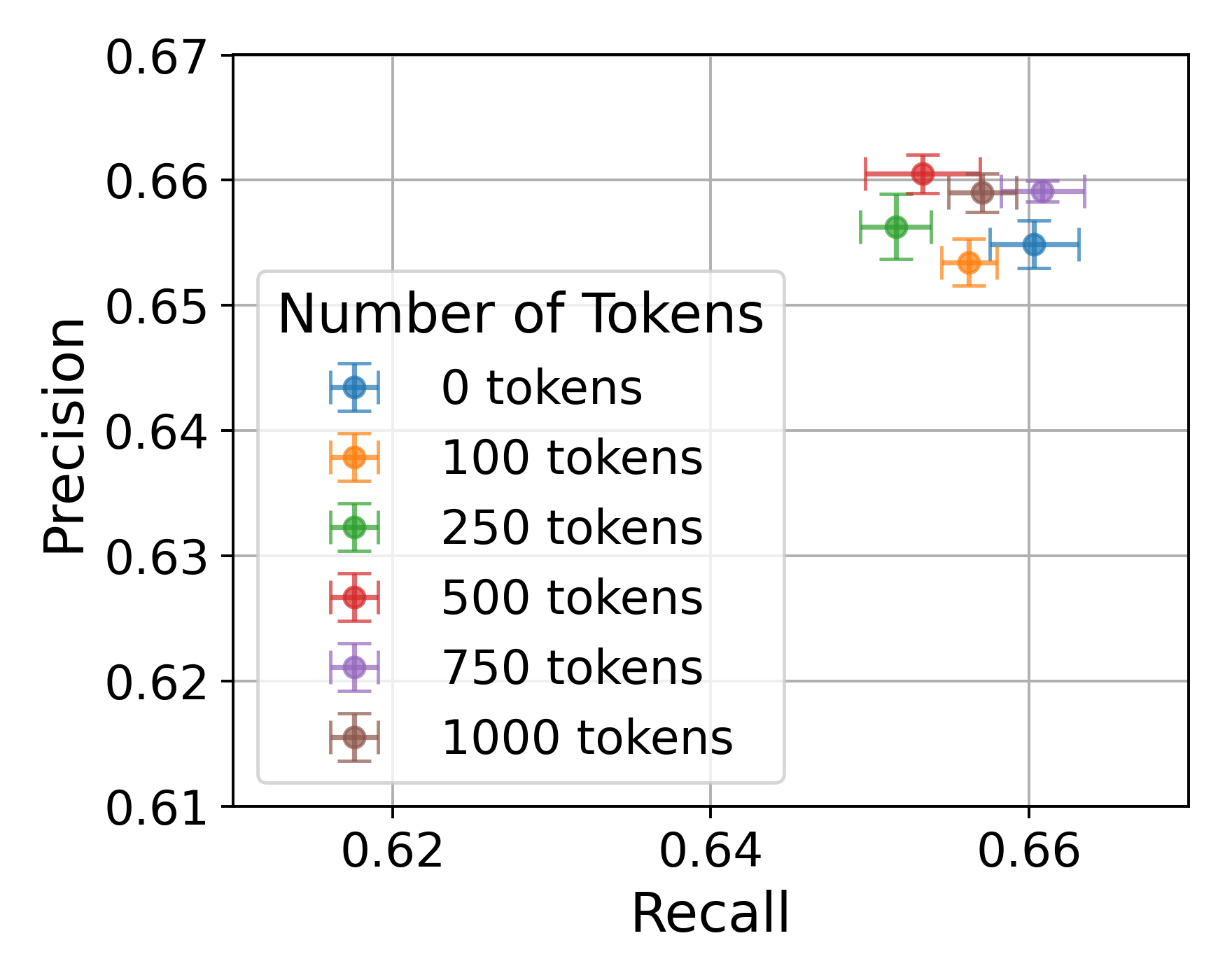}
        \caption{$\mathsf{maxN}=2$}
        % \label{fig:___}
    \end{subfigure}
    \hfill
    \begin{subfigure}[b]{0.32\textwidth}
        \centering
        \includegraphics[width=\textwidth]{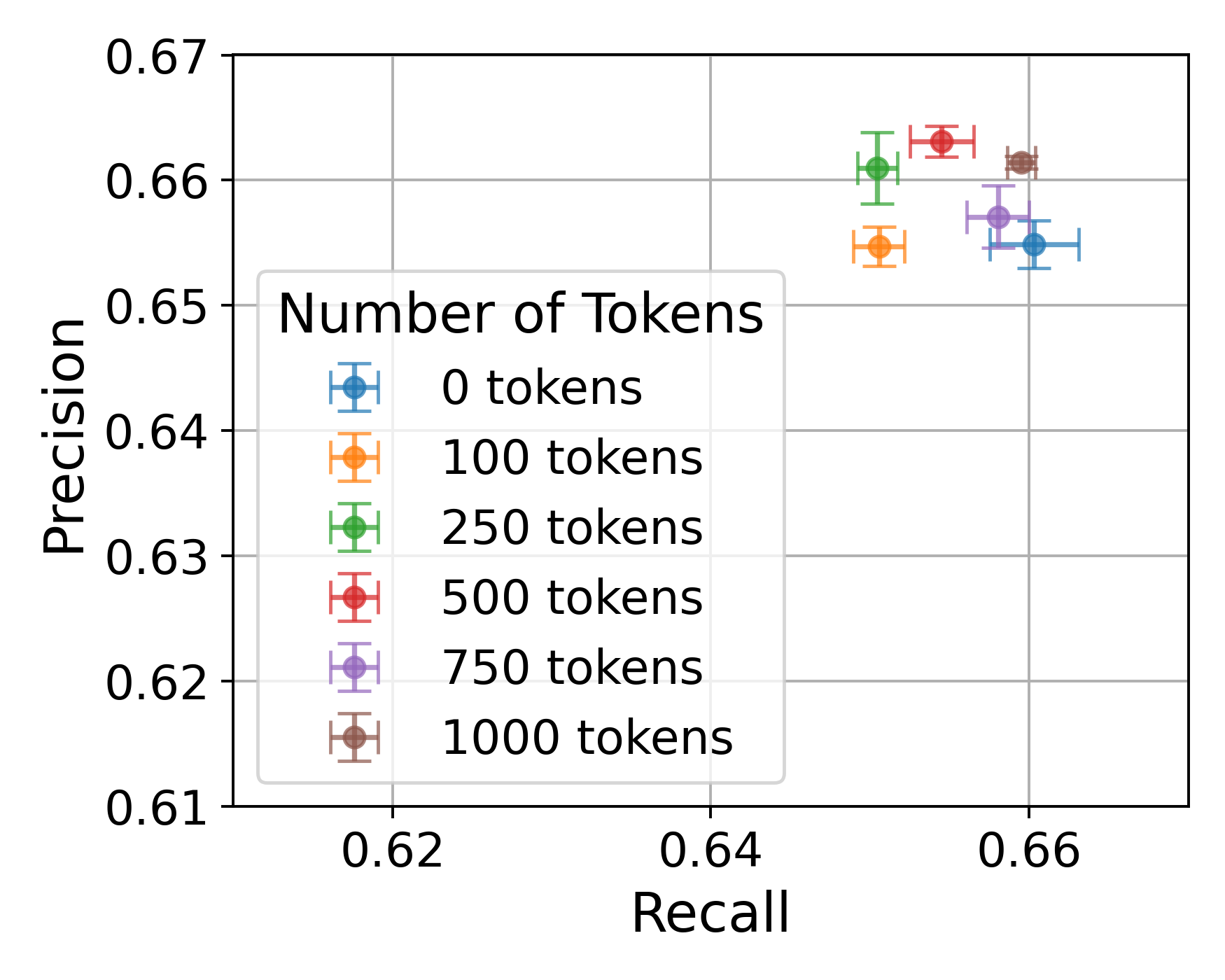}
        \caption{$\mathsf{maxN}=3$}
        % \label{fig:__}
    \end{subfigure}
    \hfill
    \begin{subfigure}[b]{0.32\textwidth}
        \centering
        \includegraphics[width=\textwidth]{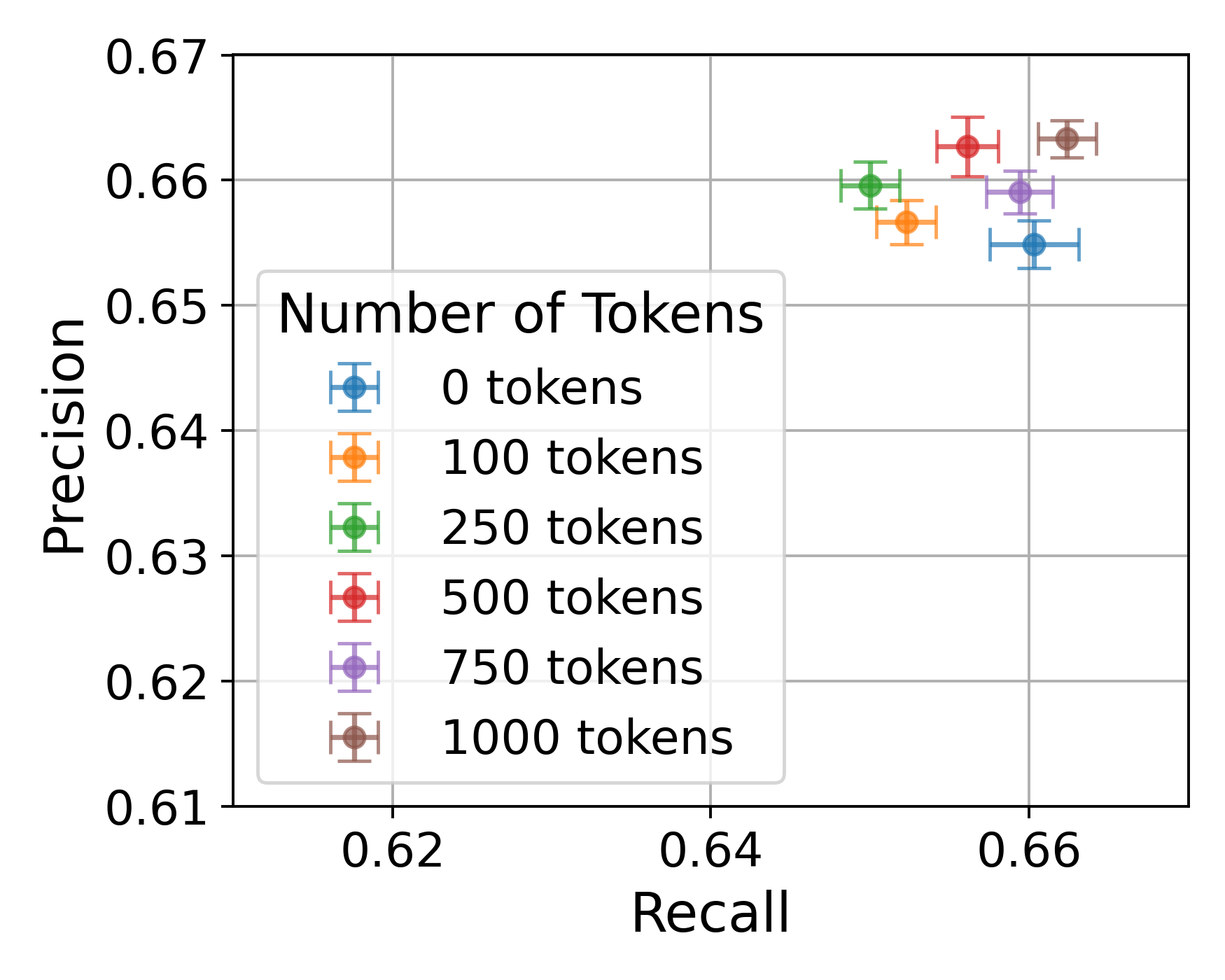}
        \caption{$\mathsf{maxN}=4$}
        % \label{fig:_}
    \end{subfigure}
    \caption{Precision-recall trade off for the \text{\eurlex} task. Each marker denotes a different number of newly added token $n$-grams to the tokenizer vocabulary prior to fine-tuning.}
    \label{fig:eurlex_pr}
\end{figure*}

\section{Additional Experiment Results and Ablations}\label{appdx:more_results}

\subsection{$\nametok$: Additional Evaluation Metrics}

\textbf{Additional Metrics in \text{\massive} Dataset:} We complement $\nametok$'s results on the ICSF task from Section \ref{sec:results} with additional metrics. Specifically, we measure the intent accuracy and slot prediction F1 score, to present a finer granularity of the fine-tuned model's behavior on the test set. For completeness, we present these metrics alongside the metrics reported in Section \ref{sec:results}. As seen from Table \ref{tab:icsf_more}, both intent accuracy and slot F1 scores present a slight improvement when introducing new $n$-gram tokens. This trend is in line with the joint accuracy, which is a coarser and stricter measure of performance.

\begin{table*}[!t]
\small
    \centering
    \begin{tabular}{||c | c c c c||}
        \toprule
        Number of tokens & Intent accuracy ($\uparrow$) &  Slot F1 ($\uparrow$) & Joint accuracy ($\uparrow$) & Average sample runtime [ms] ($\downarrow$)\\
        \midrule
        0 & 90.4 & 76.45 & 63.0 & 302 \\ 
        \midrule
        250 & \textbf{91.5}  & \textbf{77.0}  & \textbf{64.6} & 260 \\ 
        500 & 91.3 & 77.0 &  64.3 & 258.9 \\ 
        1000 & 91.1 &  76.2 & 63.4 & \textbf{256} \\
        \bottomrule
    \end{tabular}
    \caption{$\nametok$ results on the intent classification and slot fill \text{\massive} \cite{fitzgerald2023massive}  dataset.}
    \label{tab:icsf_more}
\end{table*}

\textbf{Precision-Recall:}
We present the fine-tuned model's precision-recall trade off. As the main parameter of our framework is the number of added tokens, we compare the trade-of on models trained with different numbers of added tokens. We recall that the latency improvement was already established and studied in Section \ref{sec:results}.
Figure \ref{fig:pr_icsf} presents the precision-recall trade-off for two values of number of training epochs. 
As seen from Figure \ref{fig:pr_icsf}, introducing new tokens improves precision-recall trade-off, with the best performance at $M=250$ new tokens.
These results are aligned with Table \ref{tab:icsf_more}, which shows that task performance improvement after introducing new tokens, both in terms of intent and slot fill prediction.
To that end, while existing literature often discusses a hard trade-off between runtime acceleration and output quality, we observe that this is not necessarily the case in low output variability generation with SLMs.

Next, we consider the \text{\eurlex} dataset. We observe the precision-recall trade-off for several values of $\mathsf{maxN}$, the maximal merge size in the $\nametok$ algorithm (Algorithm \ref{alg:tokenizer}). As seen in Figure \ref{fig:eurlex_pr}, the number of added tokens has a relatively negligible on the trade-off. However, we note a uniform increase in precision across all experiments.

\subsection{Comparing $\nametok$ and $\namedraft$}
This section presents a comparison the effect of $\nametok$ and $\namedraft$ in the \text{\eurlex} massive classification dataset \cite{chalkidis2019large}. The baseline is a fine-tuned model without the base tokenizer. As seen in Table \ref{tab:comparing_methods}, $\nametok$ attains a better speedup of the baseline. However, it requires access to the fine-tune procedure, while $\namedraft$ is independent of such assumption. 
\begin{table}[!t]
    \centering
    \begin{tabular}{||c|c|c||}
    \toprule
        Method & $\nametok$ & $\namedraft$ \\
        \midrule
        Speedup & $\times 2.1$ & $\times 2.17$ \\
        \bottomrule
    \end{tabular}
    \caption{Comparison of $\nametok$ and $\namedraft$.}
    \label{tab:comparing_methods}
\end{table}

\subsection{$\nametok$: Ablation on Number of Training Epochs}
We experimented with running the fine-tune procedure for different values of training epochs. It can be seen from Table \ref{tab:epoch_ablation} that a single training epoch results in insufficient performance, while training for too many epochs results in performance saturation, and even decrease in some cases (which may follow from overfit). To that end, we chose to conduct experiments with $3$ training epochs, which demonstrated satisfactory performance, while providing us with a relatively lightweight fine-tune procedure.

\begin{table}[!ht]
    \centering
    \begin{tabular}{||c|c|c||}
    \toprule
    \multicolumn{3}{||c||}{\textbf{EUR-LEX}} \\
    \midrule
        \# Train Epochs & F1 & Avg. runtime  \\
        \midrule
        $1$ & $0.624$ & $0.174$\\
        $3$ & $0.647$ & $0.174$\\
        $5$ & $\mathbf{0.655}$ & $\mathbf{0.173}$\\
        $10$ & $0.637$ & $0.181$\\
        \midrule
        \multicolumn{3}{||c||}{\textbf{MASSIVE}} \\
        \midrule
        \# Train Epochs & Joint acc. & Avg. runtime \\
        \midrule
        $1$ & $0.576$ & $0.254$ \\
        $3$ & $\mathbf{0.634}$ & $\mathbf{0.256}$ \\
        $5$ & $0.625$ & $0.257$ \\
        \bottomrule
    \end{tabular}

    \caption{Number of training Epochs ablation, number of added tokens is $M=1000$ across all runs.}
    \label{tab:epoch_ablation}
\end{table}

\subsection{$\nametok$: Results on Additional Models}
We present additional results on the $\nametok$ method by varying the utilized model. We consider the Mistralv02-7B model and apply it to the massive classification task. As seen from table \ref{tab:mistral_eurlex}, the F1 score is maintained after introducing new tokens to the LLM's tokenizer while exhibiting a decrease in the average runtime on the test data. However, we note that the obtained efficiency gains are slightly smaller than the ones obtained for the Qwen2.5-0.5B model. Specifically, Mistral obtains a speedup of $\times 1.19$, while Qwen obtained a speedup of $\times 2.31$.

\begin{table}[]
\small
    \centering
    \begin{tabular}{||c|c| c||}
    \toprule
       \# Added Tokens  & F1 score & Average Runtime [ms]\\
       \midrule
       0  &   $69.1 \pm 8$ & $1280 \pm 70.1$ \\
       150   &   $68.9 \pm 0.09$ & $1220 \pm 81.3$ \\
        500 &   $68.3 \pm 0.5$ & $1190 \pm 79.4$ \\
        1000 & $ 68.8 \pm 0.5$ & $1130 \pm 90.1$\\
        \bottomrule
    \end{tabular}
    \caption{$\nametok$ Results on \text{\eurlex} dataset with model MistralV02-7B.}
    \label{tab:mistral_eurlex}
\end{table}

\subsection{$\namedraft$: Ablation on Mixing Parameter Value}
\begin{figure*}[!t]
    \centering
    \begin{subfigure}[b]{0.46\textwidth}
        \centering
        \includegraphics[width=\textwidth]{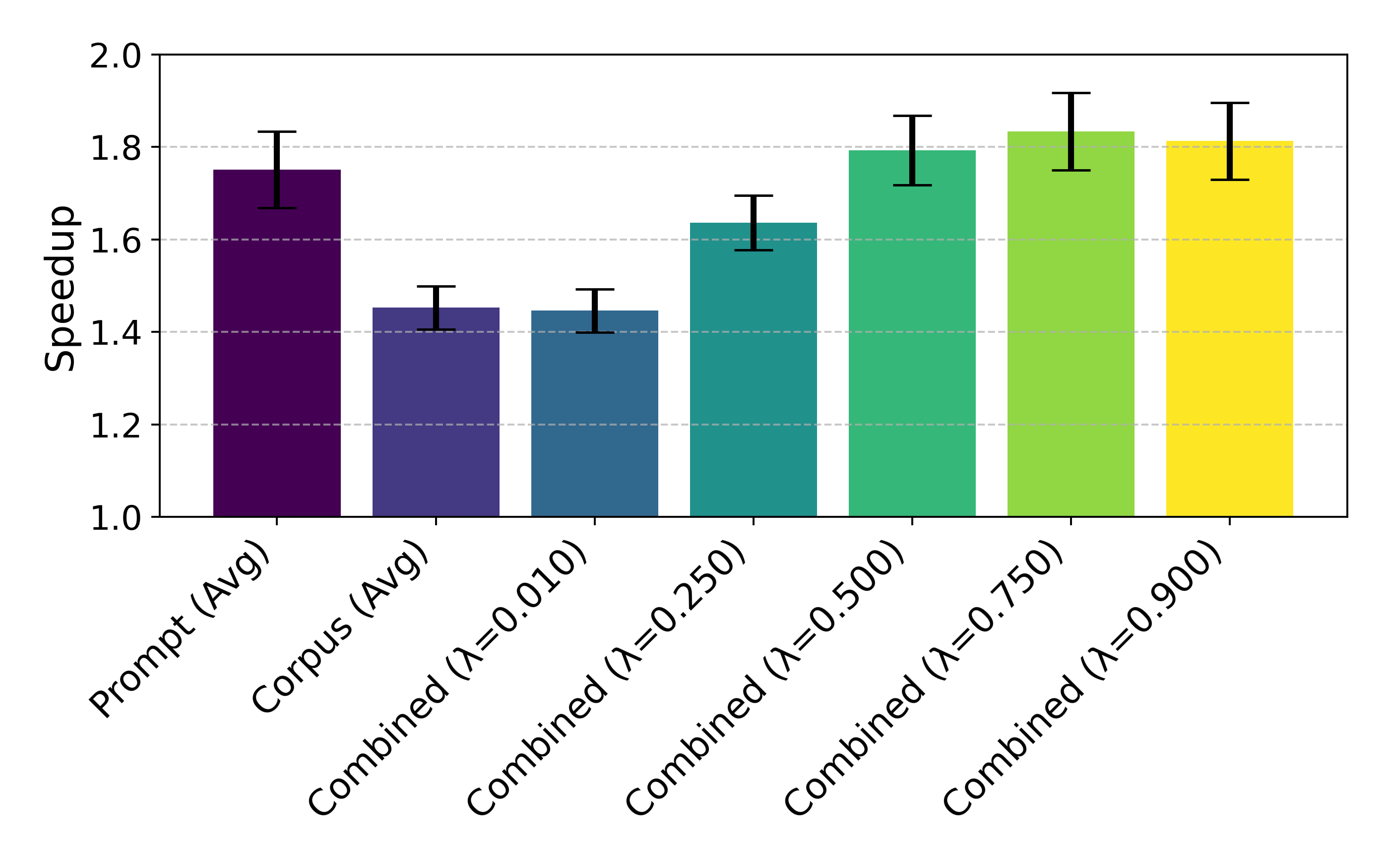}
        \caption{MedQuaD dataset}
        % \label{fig:___}
    \end{subfigure}
    \hfill
    \begin{subfigure}[b]{0.46\textwidth}
        \centering
        \includegraphics[width=\textwidth]{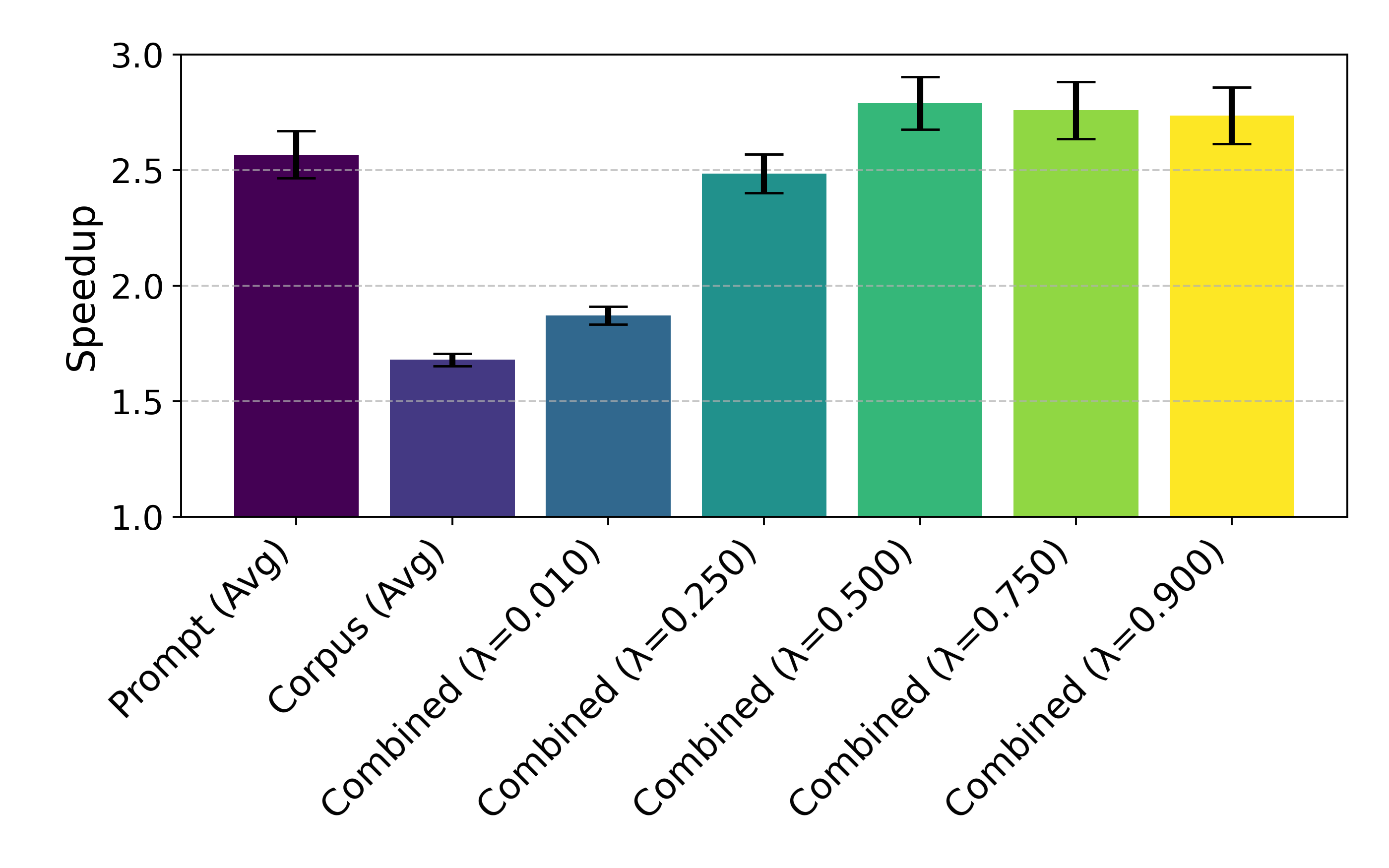}
        \caption{\text{\clinical} dataset.}
        % \label{fig:__}
    \end{subfigure}
    
    \caption{Speedup vs. mixing parameter $\gamma$.}
    \label{fig:speedup_vs_gamma}
\end{figure*}

We test the effect of the mixing parameter $\lambda$ value in $\namedraft$. 
To do that, we curate a dedicated validation set of $150$ samples from each dataset.
As seen in Figure \ref{fig:speedup_vs_gamma}, For $\lambda\in[0.5,0.9]$, we result in a mixture distribution that improves upon both the corpus $n$-gram and the prompt information. We note that highest speedup is attained within the interval $\lambda\in[0.5,0.75]$, which implies that the underlying distributions are relatively spiky. That is, each distribution proposes a next token with relatively high probability in distinct cases, resulting in more accepted token overall. 
We note that unlike n-grammys \cite{stewart2024n}, the $n$-gram distributions are independent of the underlying model, and that potential misalignment with the model might decrease performance, while the $n$-gram model properly predict the "task ground truth" next token.

\begin{figure*}[!t]
    \centering
    \begin{subfigure}[b]{0.45\textwidth}
        \centering
        \includegraphics[width=\textwidth]{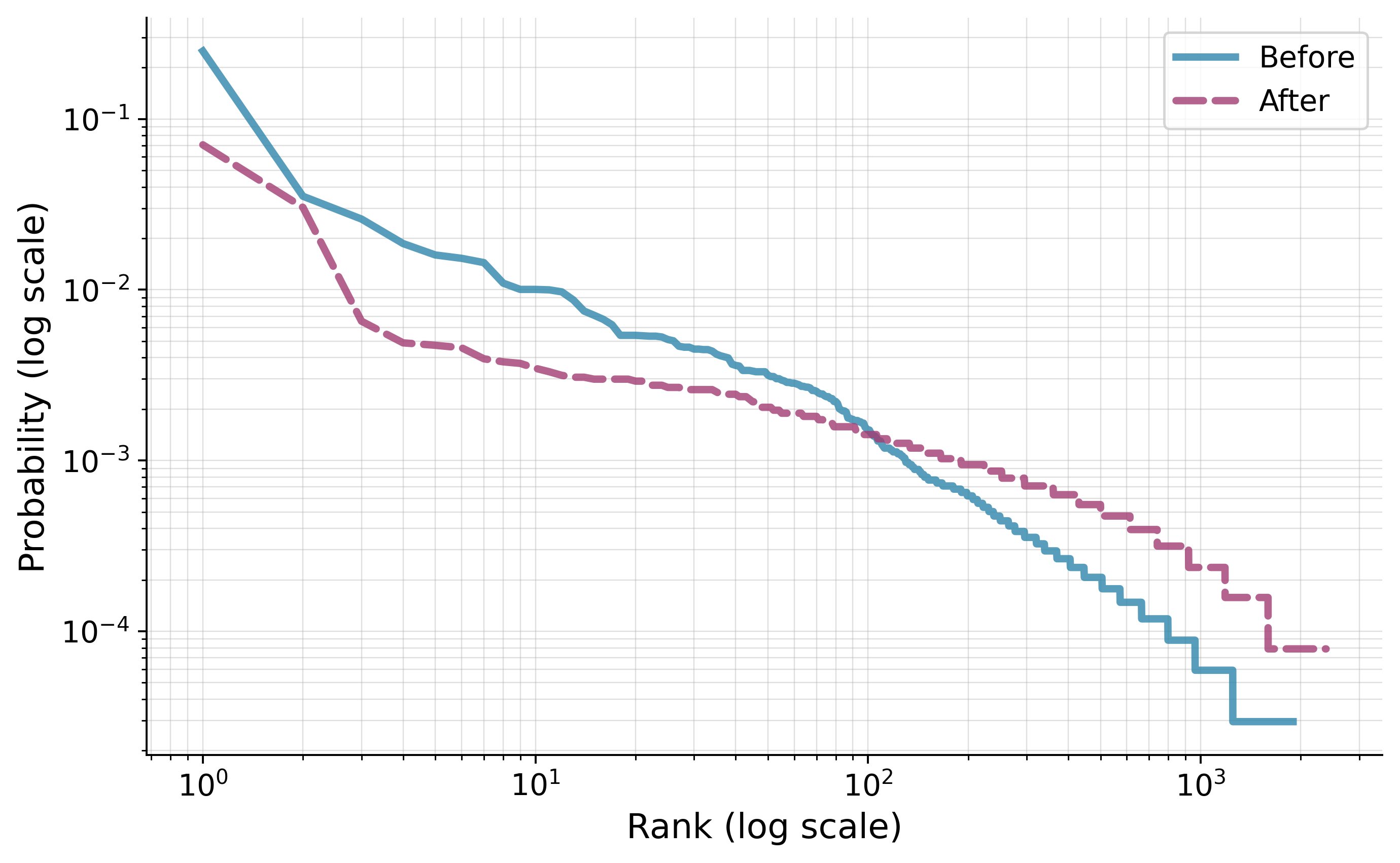}
        \caption{\text{\eurlex} dataset (massive multi-class multi-label calssification).}
        % \label{fig:___}
    \end{subfigure}
    \hfill
    \begin{subfigure}[b]{0.45\textwidth}
        \centering
        \includegraphics[width=\textwidth]{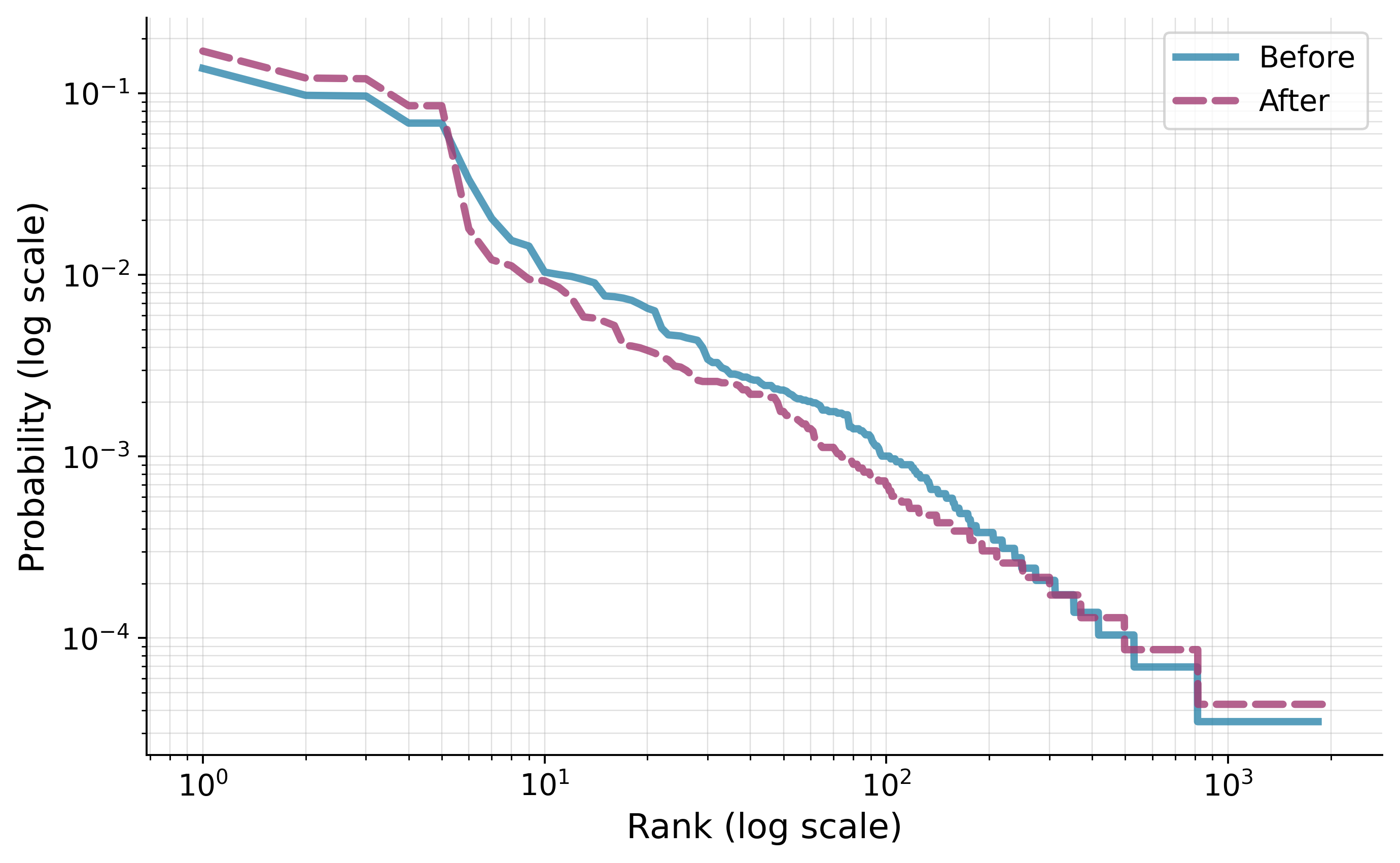}
        \caption{\text{\massive} dataset (intent classification \& slot fill).}
        % \label{fig:__}
    \end{subfigure}
    \hfill
    \begin{subfigure}[b]{0.45\textwidth}
        \centering
        \includegraphics[width=\textwidth]{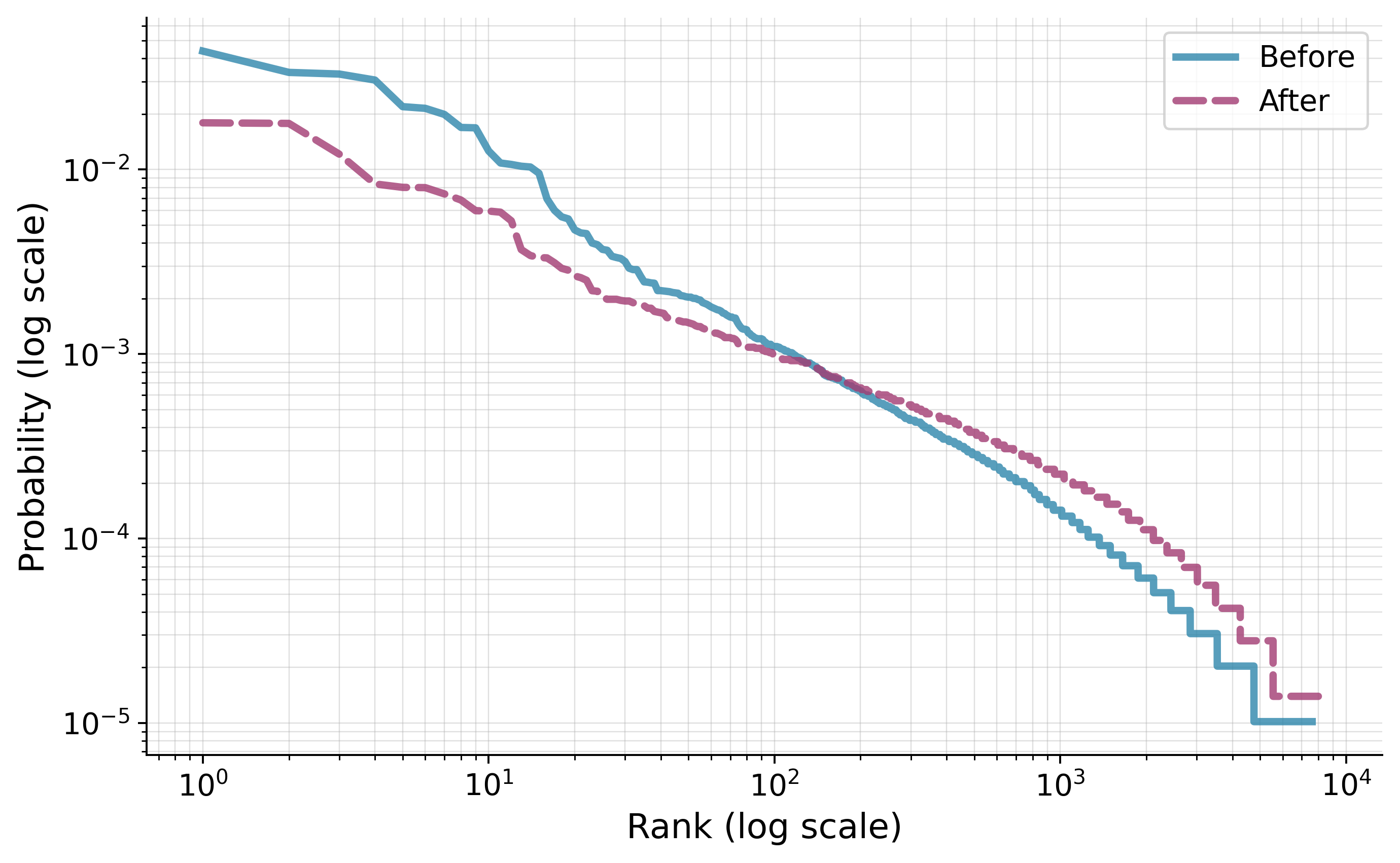}
        \caption{\text{\clinical} dataset (medical QA).}
        % \label{fig:_}
    \end{subfigure}
    \caption{Token distribution comparison before and after augmenting the tokenizer vocabulary via $\nametok$ (Algorithm \ref{alg:tokenizer}). I cane be seen that applying $\nametok$ results in a more uniform token distribution.}
    \label{fig:uid}
\end{figure*}

\section{Uniform Information Density - Analysis}\label{appdx:uid}
We provide empirical evidence that the effects induced by $\nametok$ are consistent with the uniform information distribution (UID) hypothesis.
UID predicts that efficient representations distribute information more evenly across tokens, reducing concentration of probability mass and stabilizing per-token surprisal along sequences.
By collapsing frequent task-specific $n$-grams into atomic symbols, $\nametok$ increases the size of the \textit{token} larger typical set.
As a result, the empirical output token distribution becomes less peaked and exhibits a heavier tail.
This effect is visualized in Figure~\ref{fig:uid}, where applying $\nametok$ leads to a noticeably flatter distribution compared to the baseline tokenizer.

We further quantify this uniformization by measuring the entropy of the output token distribution before and after applying $\nametok$.
We report normalized entropy, obtained by dividing the empirical entropy by the maximal possible entropy, $\log |\mathcal{V}|$.
As shown in Table~\ref{tab:token_ent_nametok}, $\nametok$ consistently increases token entropy across settings, indicating reduced redundancy.

\begin{table*}[!t]
\small
    \centering
    \begin{tabular}{||c||c|c|c|c||}
        \toprule
         - & \multicolumn{2}{|c|}{Output}& \multicolumn{2}{|c||}{Input} \\ 
         \midrule 
        Dataset & Entropy (before) &  Entropy (after) & Entropy (before) & Entropy (after)\\
        \midrule
        \text{\eurlex} & 0.423 & 0.592 & 0.4502 & 0.455 \\ 
        \midrule
        \text{\clinical} & 0.402  & 0.537  & 0.479 & 0.562 \\ 
        % 500 & 0.913 & 0.770 &  0.643 & 0.2589 \\ 
        % 1000 & 0.911 &  0.762 & 0.634 & \textbf{0.256} \\
        \bottomrule
    \end{tabular}
    \caption{Effect of $\nametok$ on the token entropy in the output dataset.}
    \label{tab:token_ent_nametok}
\end{table*}

\section{Predicting Model Speedup in $\nametok$ from the Data}
\label{app:proxy_validation}
Our goal is to validate whether a cheap preprocessing-time statistic such as the R\'enyi entropy of order 2 ($H_2$) can serve as a reliable proxy for predicting runtime reduction after fine-tuning. Following \cite{zouhar2023tokenization}, R\'enyi entropy of the token distribution was shown to be correlated with model performance. Furthermore, we consider entropic measures are a proxy for the size of the typical set due to \cite{cover1999elements} (See Section \ref{sec:output_ngrams}). For a distribution $p(x)$ defined on some finite set $\cX$, The R\'enyi entropy of order $\alpha\neq1$ is defined as
$$
H_\alpha(p) = \frac{1}{1-\alpha}\log\left(\sum_{x\in\cX} p(x)^\alpha\right),
$$
where $\lim_{\alpha\to 1}H_\alpha(p)$ converges to Shannon's entropy.

\textbf{Setting:} We consider a total of $124$ simulations, spanning over two datasets (\text{\eurlex} and \text{\clinical}), two SLM sizes (Qwen2.5-0.5B and Qwen2.5-3B), $3$ values of $n_{\mathsf{max}}$ ($2,3,4$) and several seed values. Each simluation is obtained as follows: We set a token budget $M$, apply Algorithm \ref{alg:tokenizer} to collect $M$ tokens, measure the augmented token vocabulary R\'enyi entropy (over the train set), fine-tune the model with the augmented tokenizer and record the average runtime on the test set.
We group the samples as sequences of $(M, H_2(M), \mathsf{runtime}(M))$ for each set of hyperparameter values.
\textbf{Our goal is to analyze the correlation of predictive power of $H_2$ for fine-tuned model runtime reduction}.

\subsection{Ordering Consistency via Rank Correlation}
To assess the predictive power of $H_2$, we compute Kendall’s rank correlation coefficient $\tau$ between $H_2(M)$ and $\mathsf{runtime}(M)$ for each sequence.
Kendall’s $\tau$ measures how consistently two quantities induce the same ordering over a set of items.
A higher value of $\tau$ indicates that pairs of items that are ranked higher by one metric are also more likely to be ranked higher by the other, reflecting stronger agreement in their relative ordering.
Conversely, values of $\tau$ close to zero indicate little systematic agreement beyond chance, while negative values indicate that the two metrics tend to induce opposing orderings.
We compute $\tau$ separately for each sequence and report aggregate statistics across sequences, thereby assessing robustness to dataset and hyperparameter variation.
Across all $124$ simulations, divided into $8$ settings we observe a mean Kendall’s $\tau$ of $-0.964$, indicating strong and consistent ordering agreement between $H_2(M)$ and $\mathsf{runtime}(M)$.

\subsection{Directional Predictive Reliability}
While rank correlation captures global ordering agreement, it does not directly address the operational question of interest:
\emph{does improving the $H_2(M)$ reliably lead to an improvement in $\mathsf{runtime}(M)$?} \cite{pesaran1992simple}
To answer this, we analyze parameter intervals updates each sequence.
For consecutive values $M_i, M_{i+1}$, we compute
\begin{align*}
    &\Delta H_2 = H_2(M_{i+1}) - H_2(M_i), \\
    &\Delta R = \mathrm{runtime}(M_{i+1}) - \mathsf{runtime}(M_i).
\end{align*}

Conditioning on updates where the R\'enyi entropy decreases ($\Delta(H_2) > 0$), we measure the conditional \textit{success} rate
$$
\Pr\left(\Delta R < 0 | \Delta(H_2) > 0\right),
$$
i.e., the probability that runtime decreases despite given an increase in the R\'enyi entropy.
Its complement is defined as a directional success rate.
Based on the data collected from $124$ simulations, the directional success rate is $92\%$, implying that increase in $H_2$ are followed by runtime reductions in the vast majority of cases.

\begin{figure*}[!b]
\centering
\begin{systemprompt}
\ttfamily
You are an intelligent clinical language model.

[Discharge Summary Begin] 
\newline
\{Discharge Summary\} 
\newline
[Discharge Summary End]

[Instruction Begin] 
\{Instruction\}
[Instruction End]

[Agent A's Answer Begin]
\newline
\{A\} 
\newline
[Agent A's Answer End]
\newline
\newline
[Agent B's Answer Begin] 
\newline
\{B\} 
\newline
[Agent B's Answer End] 
\newline
\newline

Above, we provide you with a discharge summary
and the instruction that the healthcare professional
gave about the discharge summary. You are also
provided with \{Number of Samples\} corresponding responses from \{Number of Samples\} different
clinical models. Your task is to read the discharge
summary and the instruction carefully, then find the
answer to the instruction. Then, compare your answer
with each model's response and evaluate the
response based on the following criteria.

Criteria:
\begin{enumerate}
\item Unacceptable (1 point): The model's response
includes any incorrect or irrelevant content. If the
instruction was unanswerable, the model did not
acknowledge this and outputs the wrong answer.
\item Poor (2 points): The model's response does not
contain any incorrect or irrelevant content but omits
significant or crucial content required by the instruction.
\item Satisfactory (3 points): The model's response
does not contain any incorrect or irrelevant content
but omits minor or insignificant required content.
\item Excellent (4 points): The model's response
contains all necessary information. If the instruction
was unanswerable, the model correctly acknowledged
this.
\end{enumerate}

When evaluating each score, ensure that judgments
are not affected by other model responses.

The first line must contain only \{Number of Samples\}
values, separated by spaces. Output scores without
explanation.
\end{systemprompt}
\caption{Evaluation Judge LLM prompt for \text{\clinical} dataset responses (Figure \ref{tab:judge_llm_asclepius})}
\label{judge_propt}
\end{figure*}

\end{document}